\documentclass[11pt]{article}

\usepackage{acl}

\usepackage{times}
\usepackage{latexsym}

\usepackage[T1]{fontenc}

\usepackage[utf8]{inputenc}

\usepackage{microtype}

\usepackage{inconsolata}

\usepackage{graphicx}
\usepackage{amsmath}
\usepackage{amssymb}
\usepackage{amsfonts}
\usepackage{multirow}
\usepackage{booktabs}

\title{SANE: State Anomaly Neutralization for Stable Extreme-Context Delta-Rule Models}

\author{Qingwen Lin$^{1}$, Boyan Xu$^{1}$, Xiao Liu$^{3}$, Zhifeng Hao$^{1,2}$, Ruichu Cai$^{1}$\thanks{~~Corresponding author} \\
  \texttt{qingwen\_lin@foxmail.com}, \texttt{hpakyim@gmail.com}, \texttt{Liuxiao.in@gmail.com}, \\
  \texttt{haozhifeng@stu.edu.cn}, \texttt{cairuichu@gmail.com} \\[2mm]
  $^{1}$School of Computer Science, Guangdong University of Technology \\
  $^{2}$College of Science, Shantou University 
  $^{3}$Yuanshi Intelligence
}

\begin{document}
\maketitle

\begin{abstract}
Delta-Rule recurrent models maintain a fixed-size state, enabling $O(1)$ inference memory but potentially becoming unstable under extreme-context extrapolation.
By tracking RWKV-7 over sequences of up to 100M tokens, we empirically identify a distinct failure pattern: \textbf{localized norm explosion atop a relatively sparse substrate}, rather than global state saturation.
Analysis of the recurrent update suggests that persistent decay keeps weakly updated entries small, whereas uneven injections allow a few channels to accumulate extreme values.
Motivated by this diagnosis, we propose \textbf{State Anomaly Neutralization (SANE)}, which applies adaptive $\tanh$ compression at chunk boundaries while preserving the intra-chunk parallel structure.
Within a safe threshold range ($3 \le \alpha \le 5$), SANE matches the baseline on 11 short-context reasoning benchmarks with no statistically significant degradation. After a 100M-token prefix, which exceeds the training length by over $24{,}000\times$, SANE retains functional reasoning ($33.46$--$35.56$) while the baseline encounters numerical overflow.
In contrast, overly permissive thresholds ($\alpha \ge 8$) remain numerically stable but lose reasoning capability entirely, showing that numerical stabilization alone does not guarantee functional reasoning and revealing a capacity--stability trade-off in state compression.
\end{abstract}

\section{Introduction}
\label{sec:intro}

\begin{figure}[!t]
    \centering
    \includegraphics[width=\columnwidth]{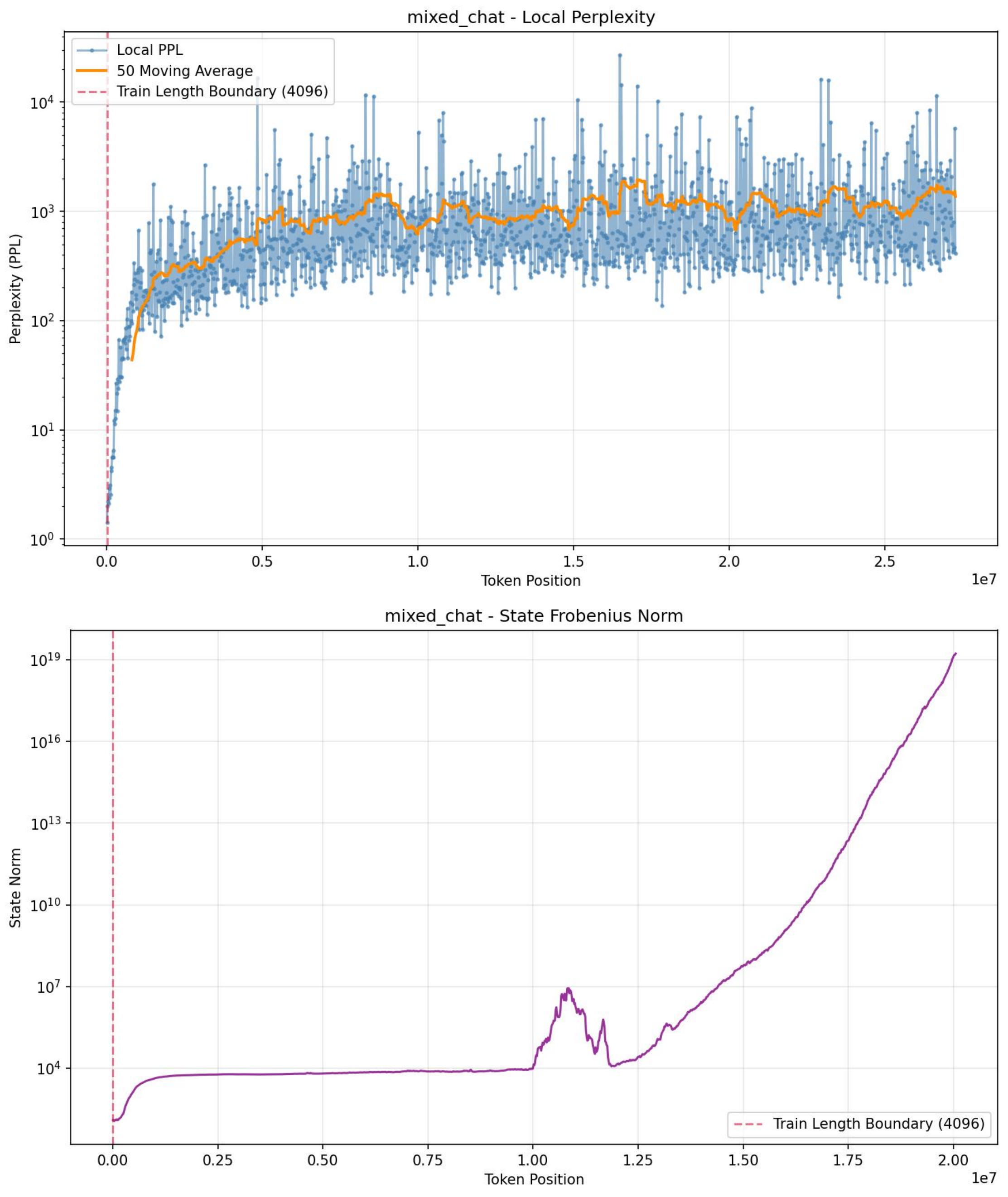}
    \caption{
Extreme-context degradation of RWKV-7 on \texttt{mixed\_chat}.
\textbf{Top:} Local perplexity and its 50-token moving average.
\textbf{Bottom:} State Frobenius norm, which eventually exhibits runaway growth beyond $10^{19}$.}
    \label{fig:long-ppl}
\end{figure}

In recent years, Delta-Rule-based recurrent models, including RWKV-7~\cite{peng2023rwkv,peng2025rwkv,peng2024eagle}, GDN~\cite{yang2026beyond,cai2026u}, and KDA~\cite{team2025kimi,zhang2026structural}, have achieved competitive sequence modeling performance in large language models, establishing an important direction within linear recurrent architectures~\cite{qu2024survey,yang2025gated}. Their shared principle is to replace the Transformer KV cache with a fixed-size recurrent state matrix $S_t$, commonly updated in the affine form $S_t=S_{t-1}M_t+K_t$. Because the state size is independent of the processed sequence length $L$, these models require $O(1)$ recurrent-state memory during inference. Moreover, unlike standard Transformers, they typically avoid explicit positional encodings and normalized softmax attention over the growing token history, and are therefore not directly subject to the same positional extrapolation and attention-distribution failures~\cite{aguilar2026does,chen2025hope,shang2025longrope2,liu2025reattention}. These properties make Delta-Rule models natural candidates for long-context extrapolation, but whether their fixed-size states remain stable at lengths orders of magnitude beyond training remains poorly understood.

This structural potential does not automatically translate into stable extreme-context behavior. As shown in Figure~\ref{fig:long-ppl}, the perplexity of RWKV-7 progressively deteriorates beyond its training context, while the state norm eventually enters a runaway growth phase and exceeds $10^{19}$ near 20M tokens, after which the model encounters numerical overflow. Increasing the training length may delay this failure, but scaling training by several orders of magnitude would incur prohibitive costs. Meanwhile, most existing extrapolation methods modify positional encodings or attention distributions~\cite{chen2025hope,shang2025longrope2,huo2026periodic,rahman2025context} and do not directly address instability in fixed-size recurrent states.

\begin{figure*}[t]
  \centering
  \includegraphics[width=\textwidth]{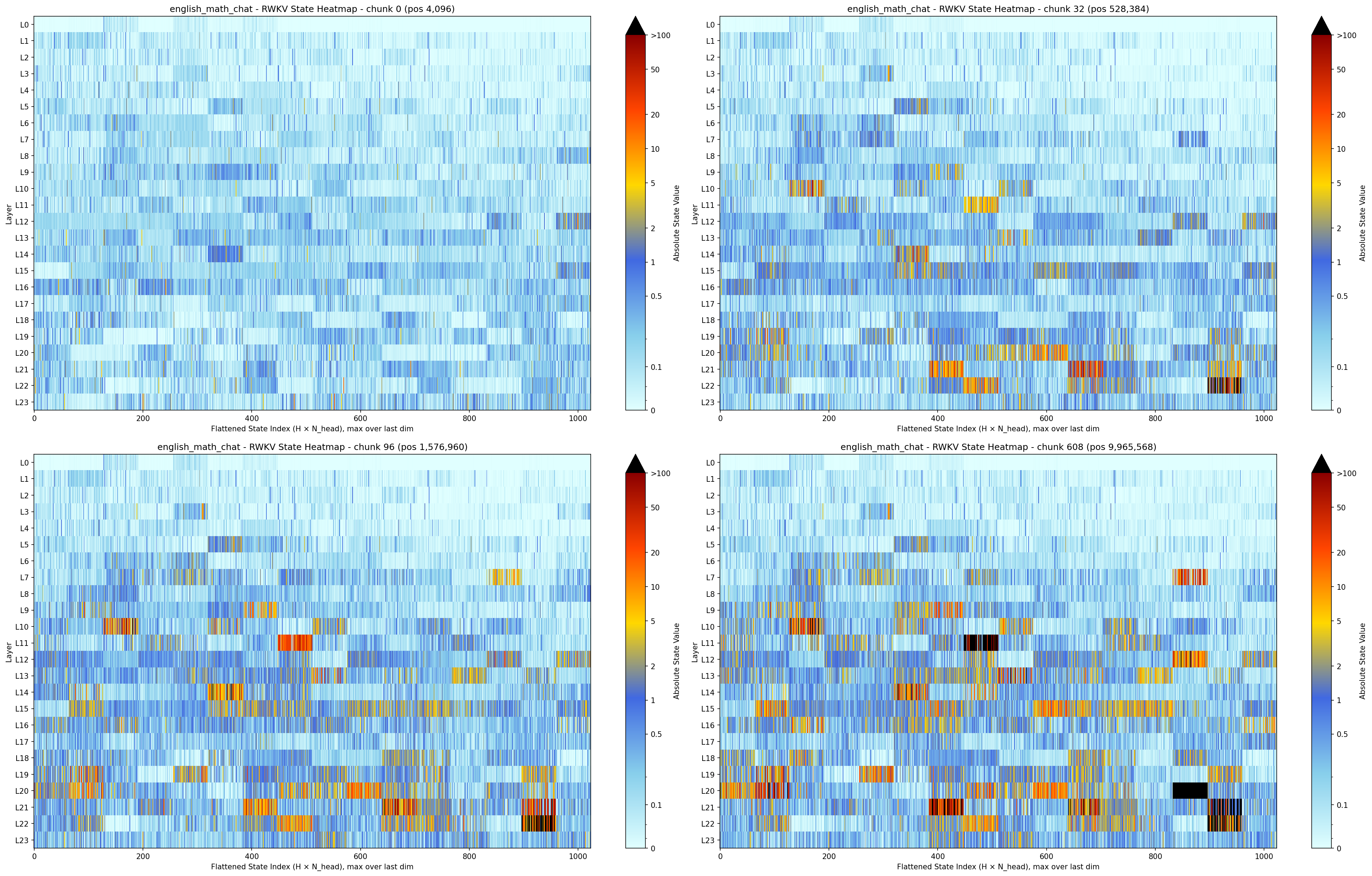}
  \caption{Spatial evolution of RWKV-7 state absolute values across increasing context lengths. 
  Each heatmap shows the absolute state values (flattened across heads and channels) 
  for all 24 layers (L0--L23). 
  \textbf{Top-left:} Within training length (4,096 tokens)---the state remains largely 
  sparse with values near zero (light blue). 
  \textbf{Top-right:} At 528K tokens---localized hotspots (yellow/orange) begin to emerge 
  in specific heads. 
  \textbf{Bottom-left:} At 1.5M tokens---the hotspots expand into concentrated red regions, 
  indicating uncontrolled growth. 
  \textbf{Bottom-right:} At 10M tokens---massive activation spreads across multiple layers, 
  with black regions marking values that exceed the display scale ($>$100).
  Crucially, the background remains sparse throughout, confirming that the collapse stems 
  from localized norm explosion rather than overall state saturation.}
  \label{fig:state-evolution}
\end{figure*}

Because the recurrent state is the principal persistent carrier of information across token positions, we systematically examine its evolution under extreme extrapolation. A finer-grained visualization in Figure~\ref{fig:state-evolution} reveals that the growth is highly non-uniform: within the training context, most state entries have magnitudes below $0.1$, while a small fraction reach approximately $10$--$20$; as the sequence grows, the low-magnitude background remains visible, whereas localized hotspots emerge and intensify by several orders of magnitude. We refer to this empirically observed magnitude separation as \emph{relative sparsity}, which describes a low-magnitude background rather than exact zeros. The expansion of these localized anomalies coincides with the escalation of the global state norm and the degradation of perplexity. Analysis of the RWKV-7 update further suggests that repeated injections and transition-induced amplification can dominate state decay in a small number of directions. We therefore characterize the observed failure mode as \textbf{localized norm explosion atop a relatively sparse substrate}, rather than global state saturation.

Motivated by this diagnosis, we propose \textbf{State Anomaly Neutralization (SANE)}, a lightweight intervention that suppresses localized anomalies while approximately preserving the low-magnitude background. SANE applies an adaptive element-wise transformation, $\widetilde{S}=\tau\tanh(S/\tau)$, to the terminal state of each chunk. The transformation is nearly identical for small $|S|/\tau$ but progressively compresses values approaching or exceeding the input-dependent threshold $\tau$. Applied only during inter-chunk state transfer, SANE prevents extreme values from being repeatedly carried forward while leaving the intra-chunk affine recurrence and associative-scan structure unchanged. It adds fewer than $0.1\%$ parameters and imposes an adaptive element-wise cap at every chunk boundary.

We evaluate SANE on RWKV-7 (0.4B) as a proof of concept by introducing it into a pretrained checkpoint through supervised fine-tuning. Across 11 short-context mathematical reasoning benchmarks, SANE matches the baseline with no statistically significant difference for $\alpha \ge 3$ ($43.30$--$44.23$ vs.\ $43.88$, all $p \ge 0.154$), while overly aggressive compression at $\alpha \le 2$ impairs short-context performance ($p \le 0.006$). After a 100M-token prefix---over $24{,}000\times$ the training context length---the baseline encounters numerical overflow, whereas SANE's average accuracy decreases monotonically with increasing $\alpha$ (Figure~\ref{fig:scaling-analysis}): within the safe regime ($\alpha \le 5$) it retains functional reasoning ($33.46$ at $\alpha{=}5$), followed by a steep decline at $\alpha{=}6$--$7$ ($18.45$ at $\alpha{=}7$) and a complete collapse to near zero for $\alpha \ge 8$. In contrast, configurations with $\alpha \ge 8$ remain numerically stable but lose reasoning capability entirely, showing that numerical stabilization alone is insufficient and revealing a capacity--stability trade-off in state compression.
In summary, the main contributions of this paper are as follows:

\begin{itemize}
\item We diagnose the state evolution of RWKV-7 under extreme extrapolation of up to 100M tokens and empirically identify \textbf{localized norm explosion atop a relatively sparse substrate}, together with a structural interpretation based on recurrent injection, amplification, and decay.

\item We propose \textbf{State Anomaly Neutralization (SANE)}, an adaptive chunk-boundary intervention that suppresses extreme state values while preserving the intra-chunk affine recurrence and parallel structure with fewer than $0.1\%$ additional parameters.

\item We show that SANE preserves short-context reasoning and retains functional accuracy after a 100M-token prefix, while contrasting threshold scales reveal that numerical stability does not necessarily imply functional health.

\end{itemize}

\section{Preliminaries}
\label{sec:preliminary}

This section introduces the affine recurrence and chunkwise computation shared by a broad class of Delta-Rule models. We then instantiate this formulation with RWKV-7 and explain why nonlinear transformations can be inserted at chunk boundaries without altering intra-chunk parallelism.

\subsection{Affine Delta-Rule Recurrence and Chunkwise Form}
\label{sec:delta-rule-def}

A broad class of Delta-Rule models maintains a fixed-size state matrix $S_t\in\mathbb{R}^{d\times d}$ whose update is affine in the previous state:
\begin{equation}
S_t=S_{t-1}M_t+K_t,
\label{eq:unified}
\end{equation}
where $M_t\in\mathbb{R}^{d\times d}$ governs the state transition and $K_t\in\mathbb{R}^{d\times d}$ injects information from the current token. This formulation originates from Delta-Net~\cite{schlag2021linear}, with subsequent models differing primarily in how $M_t$ and $K_t$ are parameterized.

Unrolling Eq.~\eqref{eq:unified} gives
\begin{equation}
S_t=S_0\left(\prod_{j=1}^{t}M_j\right)
+\sum_{j=1}^{t}K_j\left(\prod_{l=j+1}^{t}M_l\right),
\label{eq:unroll}
\end{equation}
where the matrix products are ordered chronologically. The state therefore combines the transformed initial state with historical injections propagated through subsequent transitions.

For efficient training, modern affine Delta-Rule models commonly employ chunkwise parallel computation~\cite{yang2024parallelizing}. The composition of two consecutive affine updates, where $a$ precedes $b$ in time, is
\begin{equation}
(M_a,K_a)\otimes(M_b,K_b)
=
(M_aM_b,\;K_aM_b+K_b),
\label{eq:affine-compose}
\end{equation}
which is associative. Here and throughout the paper, products are ordered chronologically:
$\prod_{j=p}^{q}M_j:=M_pM_{p+1}\cdots M_q$. Consequently, prefix states within a chunk can be computed through an associative scan.
Specifically, consider a chunk of size $C$ with initial state $S^{(c)}$. Its $i$-th state can be written as
\begin{equation}
S_i^{(c)}=S^{(c)}P_i^{(c)}+Q_i^{(c)},
\label{eq:chunkwise}
\end{equation}
where
\begin{equation*}
\begin{aligned}
P_i^{(c)}
&=\prod_{j=1}^{i}M_j^{(c)},\\
Q_i^{(c)}
&=\sum_{j=1}^{i}K_j^{(c)}
  \left(\prod_{l=j+1}^{i}M_l^{(c)}\right).
\end{aligned}
\end{equation*}
Here, an empty product is defined as the identity matrix. Because $M_j^{(c)}$ and $K_j^{(c)}$ depend only on the inputs within the current chunk and not on $S^{(c)}$, all prefix pairs $(P_i^{(c)},Q_i^{(c)})$ can be computed through an associative scan with $O(\log C)$ parallel depth. The terminal state $S_{\mathrm{end}}^{(c)}=S_C^{(c)}$ then initializes the next chunk.

\subsection{RWKV-7 as a Concrete Instantiation}
\label{sec:rwkv7-instance}

We use RWKV-7 as the experimental instantiation of the affine recurrence. Let $z_t,b_t,v_t,k_t\in\mathbb{R}^{1\times d}$ denote vectors generated from the current token, and let $w_t\in(0,1)^d$ be a data-dependent vector-valued decay. The Generalized Delta Rule (GDR) of RWKV-7 is
\begin{equation}
S_t
=
S_{t-1}
\underbrace{
\left(
\operatorname{diag}(w_t)+z_t^\top b_t
\right)
}_{M_t}
+
\underbrace{
v_t^\top k_t
}_{K_t},
\label{eq:gdr}
\end{equation}
where $z_t^\top b_t$ is a rank-one modification to the state transition and $v_t^\top k_t$ is a rank-one state injection. RWKV-7 further parameterizes
\begin{equation}
z_t=-\hat{\kappa}_t,
\qquad
b_t=\hat{\kappa}_t\odot a_t,
\label{eq:rwkv7-parameterization}
\end{equation}
where $\hat{\kappa}_t\in\mathbb{R}^{1\times d}$ is the normalized removal key and $a_t\in(0,1)^d$ is a vector-valued in-context learning rate~\cite{peng2025rwkv}. This formulation generalizes the classical Delta Rule, which admits an online-SGD interpretation under an L2 key--value prediction loss. Therefore, RWKV-7 is a concrete instance of Eq.~\eqref{eq:unified}, with a diagonal-plus-low-rank transition matrix and a rank-one injection.

\begin{figure*}[t]
  \centering
  \includegraphics[width=\textwidth]{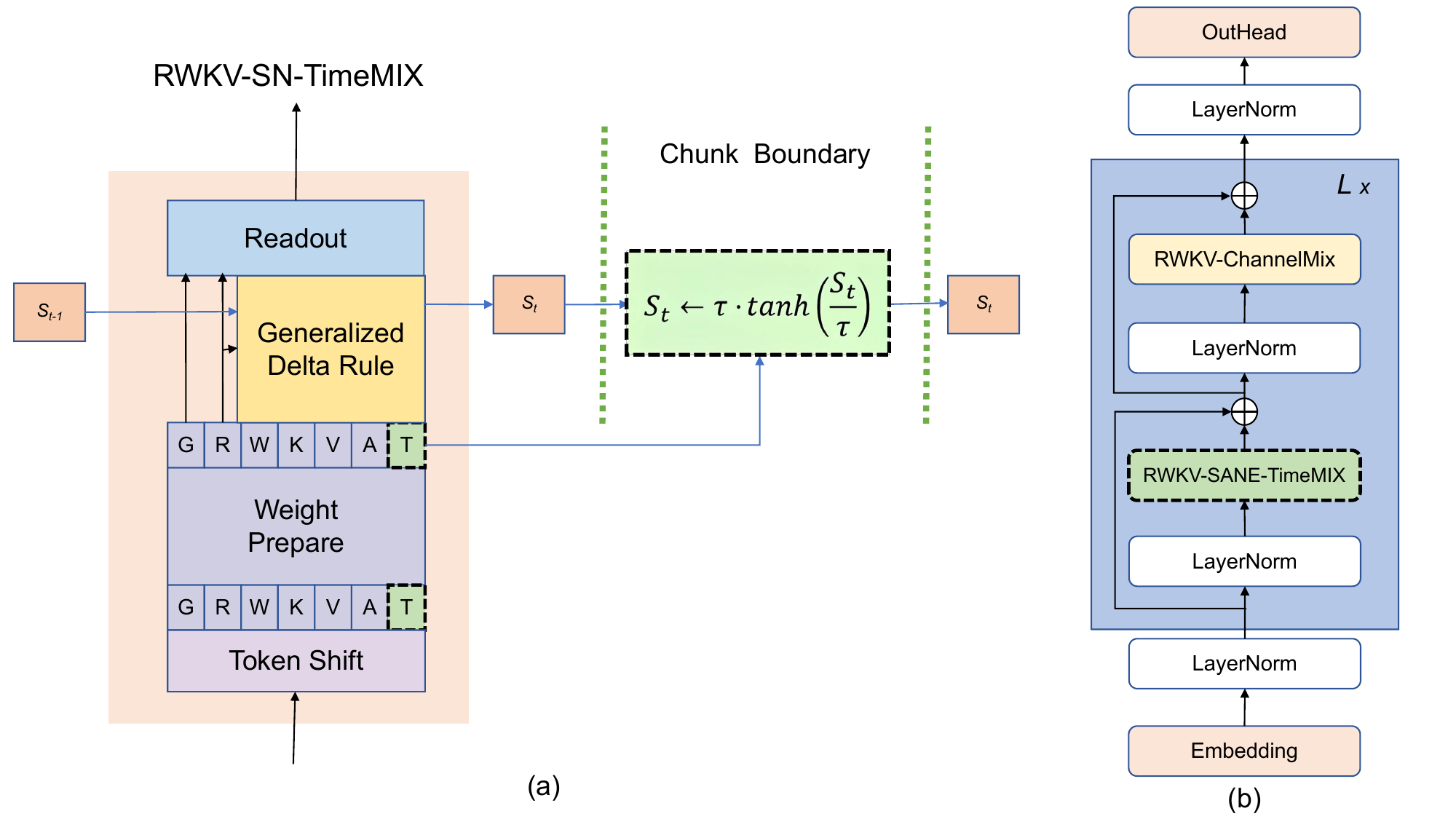}
  \caption{
    \textbf{a:} RWKV-SANE-TimeMix internals. The $\tau$ branch (green dashed) computes 
    an input-dependent threshold; State Neutralization (green dashed) is applied 
    at chunk boundaries.
    \textbf{b:} RWKV-SANE architecture. Dashed boxes mark SANE modifications.
  }
  \label{fig:architecture}
\end{figure*}

\section{Method: State Anomaly Neutralization}
\label{sec:method}

This section uses RWKV-7 as a representative generalized Delta-Rule model. We first diagnose its state evolution under extreme-context extrapolation and then interpret the observed pattern through the interaction between state decay and uneven recurrent updates. Based on this empirical diagnosis, we introduce State Anomaly Neutralization (SANE) and discuss its computational cost and compatibility with chunkwise execution.

\subsection{Extreme-Context State Diagnosis}
\label{sec:diagnosis}
We examine the state evolution of an RWKV-7-0.4B model trained with a context length of $4{,}096$. Figure~\ref{fig:long-ppl} presents the global evolution of perplexity and state norm, while Figure~\ref{fig:state-evolution} visualizes the spatial distribution of state magnitudes from $4$K to $10$M tokens.

\paragraph{Global state instability.}
As shown in the bottom panel of Figure~\ref{fig:long-ppl}, the state Frobenius norm remains within a moderate range during the early stage of extrapolation, undergoes transient fluctuations, and eventually enters a runaway growth phase. Near $20$M tokens, the norm exceeds $10^{19}$ and the evaluation encounters numerical overflow. The accompanying perplexity degradation indicates a close association between the loss of predictive quality and the escalation of state magnitude.

\paragraph{Localized growth and relative sparsity.}
Figure~\ref{fig:state-evolution} further reveals that the growth is highly nonuniform. Within the training context, most state entries remain below $0.1$, while a small fraction reach substantially larger magnitudes. We refer to this persistent order-of-magnitude separation as \emph{relative sparsity}; it describes a low-magnitude background with a small number of large entries, rather than the exact zeros induced by L1 regularization. At $528$K tokens, localized hotspots begin to appear in a few layers and heads. These hotspots intensify at $1.5$M tokens and occur in an increasing number of locations by $10$M tokens, with some values exceeding the visualization range of $100$. Nevertheless, the low-magnitude background remains visible throughout. The failure therefore differs from global state saturation and is instead characterized by localized anomalous growth.

\paragraph{Mechanistic interpretation.}
The observed pattern can be interpreted through the RWKV-7 update in Eq.~\eqref{eq:gdr}, which combines diagonal decay, a low-rank transition, and a rank-one injection. Under its online-SGD interpretation, the diagonal term plays a role analogous to per-channel weight decay and continuously contracts the inherited state, whereas the low-rank transition and injection terms update state directions unevenly. Directions receiving limited effective updates therefore tend to remain within a low-magnitude background, while frequently written directions may accumulate substantially larger values when repeated injection and transition-induced amplification dominate contraction. This interaction provides a structural explanation for the empirically observed relative sparsity and localized norm growth, but does not constitute a general sparsity guarantee.

\paragraph{Diagnosis.}
Taken together, these observations characterize the extreme-context failure of RWKV-7 as \textbf{localized norm explosion atop a relatively sparse substrate}, rather than global state saturation. This diagnosis motivates an intervention that approximately preserves the low-magnitude background while selectively suppressing extreme state values.

\subsection{State Anomaly Neutralization}
\label{subsec:sane-operation}

Motivated by the above diagnosis, we propose State Anomaly Neutralization (SANE), which applies adaptive soft compression during inter-chunk state transfer. Partition the input sequence into chunks of size $C$, and let $S^{(c)}_{\mathrm{end},h}\in\mathbb{R}^{d\times d}$ denote the terminal state of head $h$ in chunk $c$. SANE computes
\begin{equation}
\widetilde{S}^{(c)}_{\mathrm{end},h}
=
\tau^{(c)}_h
\tanh\left(
\frac{S^{(c)}_{\mathrm{end},h}}{\tau^{(c)}_h}
\right),
\label{eq:sane}
\end{equation}
where $\tau^{(c)}_h>0$ is a scalar threshold independently generated for each head.

\paragraph{Compression behavior.}
The transformation has three operating regimes. When $|S|/\tau$ is small, $\tau\tanh(S/\tau)\approx S$, so low-magnitude entries are approximately preserved; for example, the relative deviation is approximately $0.3\%$ at $|S|/\tau=0.1$. As $|S|/\tau$ increases, the transformation enters a smooth transition regime and progressively compresses the state. When $|S|/\tau\gg1$, the output approaches $\pm\tau$, suppressing extreme values before they are carried into subsequent chunks.

\paragraph{Input-dependent thresholding.}
The threshold is generated from the last-token hidden representation of the current chunk:
\begin{equation}
\tau^{(c)}_h
=
\alpha\,
\mathrm{softplus}
\left(
x^{(c)}_{\tau}W^{(h)}_{\tau}
\right)
+1,
\label{eq:tau}
\end{equation}
where $W_{\tau}\in\mathbb{R}^{d_{\mathrm{model}}\times H}$ is learnable and $\alpha$ is a preset scale controlling the overall compression strength. The softplus transformation and constant offset ensure $\tau^{(c)}_h>1$. We zero-initialize $W_{\tau}$, producing an initial threshold of approximately $0.69\alpha+1$. During training, the threshold adapts to the current chunk representation. A smaller $\alpha$ produces stronger compression, whereas a larger $\alpha$ preserves a wider range of state magnitudes.

The design of SANE directly reflects the diagnosis in Section~\ref{sec:diagnosis}. Because the empirically observed state background is concentrated at low magnitudes, most background entries fall within the near-identity regime and are only weakly perturbed. In contrast, anomalously large entries enter the transition or saturation regime and are progressively suppressed. Independent per-head thresholds allow the compression scale to vary across heads, accommodating the observed spatial nonuniformity of state growth without imposing a single global threshold.

\paragraph{Boundary-wise magnitude control.}
Because $|\tanh(z)|\leq1$, Eq.~\eqref{eq:sane} ensures that, for every realized threshold,
\begin{equation}
\left|
\widetilde{S}^{(c)}_{\mathrm{end},h,ij}
\right|
\leq
\tau^{(c)}_h,
\qquad
\left\|
\widetilde{S}^{(c)}_{\mathrm{end},h}
\right\|_F
\leq
d\,\tau^{(c)}_h.
\label{eq:boundary-control}
\end{equation}
SANE therefore directly controls the magnitude of the state transferred to the next chunk. However, because $\tau^{(c)}_h$ is input dependent and the softplus function does not impose a fixed global upper bound, Eq.~\eqref{eq:boundary-control} should be understood as adaptive boundary-wise control rather than a uniform boundary-wise magnitude control over arbitrary inputs and sequence lengths. Moreover, numerical magnitude control alone does not guarantee functional reasoning, as different threshold scales may preserve different amounts of useful state information.

\paragraph{Chunk-boundary placement.}
SANE is applied only after the affine recurrence within a chunk has been completed and before its terminal state is passed to the next chunk. Consequently, the intra-chunk recurrence and associative-scan computation remain unchanged. In our RWKV-7 implementation, we reuse the existing chunk size $C=16$, so $\alpha$ is the only additional manually selected hyperparameter. The computational and memory costs of this intervention are analyzed in Section~\ref{subsec:cost}.

\subsection{Cost and Chunkwise Compatibility}
\label{subsec:cost}

SANE introduces a lightweight modification to inter-chunk state transfer. We analyze its parameter count, computational complexity, memory requirements, and compatibility with chunkwise execution below.

\paragraph{Parameters.}
Each layer introduces the projection matrix $W_{\tau}\in\mathbb{R}^{d_{\mathrm{model}}\times H}$ and the Token Shift coefficient of the threshold branch. In our RWKV-7-0.4B implementation, these components add approximately $0.39$M parameters in total, corresponding to less than $0.1\%$ of the original model size.

\paragraph{Computation.}
At every chunk boundary, SANE computes a projection from the last-token representation to $H$ thresholds and applies the element-wise transformation in Eq.~\eqref{eq:sane} to the $H$ state matrices. The additional complexity per chunk is
\begin{equation}
O\left(d_{\mathrm{model}}H+Hd^2\right),
\end{equation}
which is amortized over $C$ tokens. The corresponding per-token overhead is therefore
\begin{equation}
O\left(\frac{d_{\mathrm{model}}H+Hd^2}{C}\right).
\end{equation}
Because the state transformation is executed once every $C$ steps rather than at every token, its computational cost is expected to be small relative to the recurrent block. We report the analytical overhead here; a detailed evaluation of end-to-end throughput and latency is left for future work.

\paragraph{Memory.}
SANE does not introduce a sequence-length-dependent cache. During serial inference, the transformed state can replace the terminal state passed to the next chunk, requiring only the $H$ threshold values and, depending on the implementation, a temporary state-sized buffer. During training, additional activations are confined to chunk boundaries. Thus, the persistent inference-memory complexity remains independent of the processed context length, although the operation should not be interpreted as having strictly zero memory overhead.

\paragraph{Chunkwise compatibility.}
SANE is applied after the affine recurrence within a chunk and before the terminal state is transferred to the next chunk. It therefore leaves the intra-chunk associative scan and its $O(\log C)$ parallel depth unchanged. The nonlinear transformation prevents multiple chunks from being merged into a single affine scan, but standard chunkwise execution already transfers terminal states between consecutive chunks. Consequently, SANE preserves intra-chunk parallelism while introducing one additional inter-chunk operation at each boundary.

Overall, SANE adds fewer than $0.1\%$ parameters in our RWKV-7-0.4B implementation, retains sequence-length-independent inference memory, and preserves the existing intra-chunk parallel structure. It should therefore be viewed as a lightweight modification to chunkwise state transfer rather than a zero-overhead or universally plug-and-play module.

\begin{figure*}[t]
  \centering
  \includegraphics[width=\textwidth]{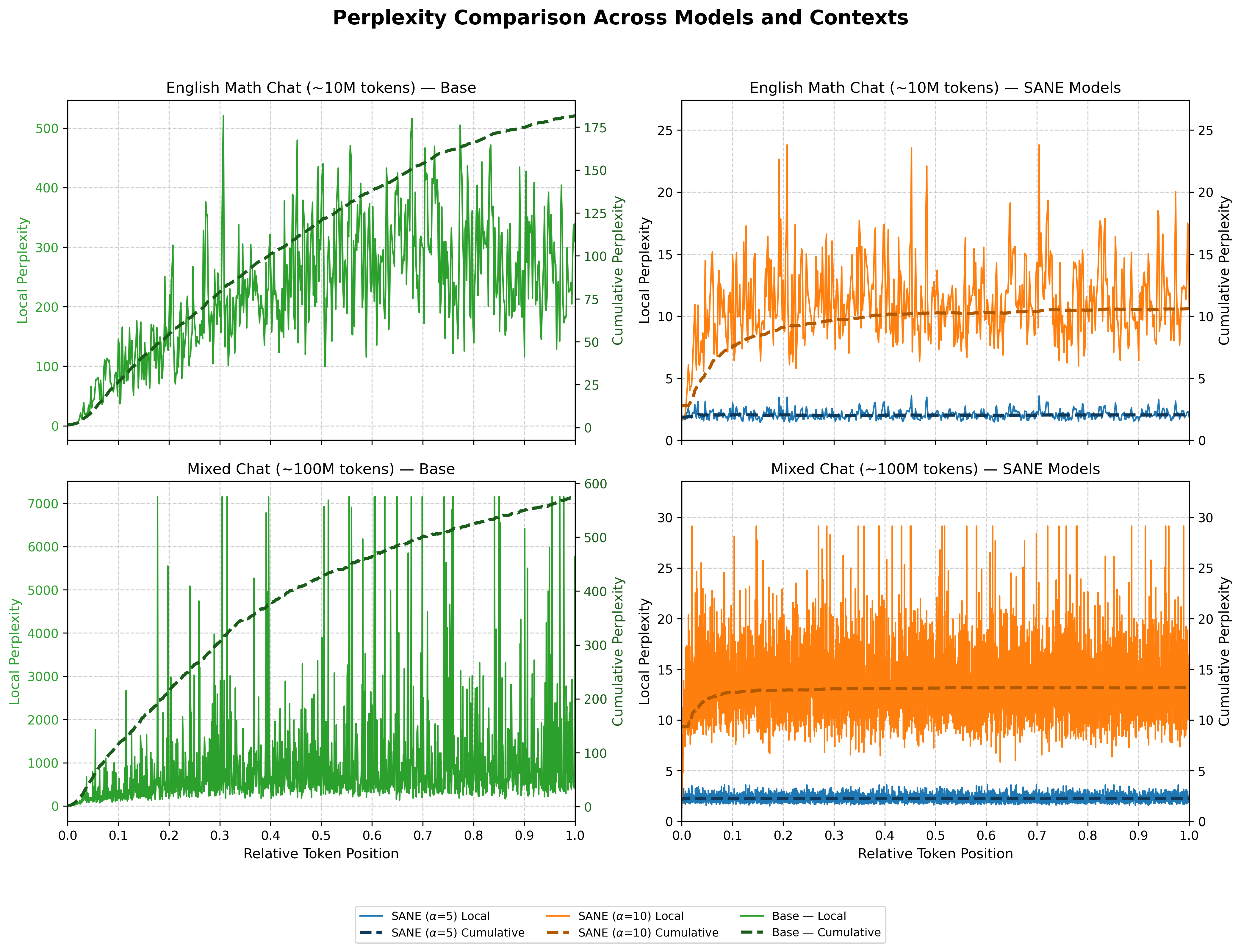}
    \caption{Perplexity evolution under extreme-length extrapolation. 
    \textbf{Left column:} Baseline (RWKV-0.4B). 
    \textbf{Right column:} SANE models. 
    \textbf{Top row:} \texttt{english\_math\_chat} ($\sim$10M tokens). 
    \textbf{Bottom row:} \texttt{mixed\_chat} ($\sim$100M tokens). 
    Solid lines denote local perplexity at each token position; 
    dashed lines denote cumulative perplexity smoothed over the prefix. 
    The baseline exhibits unbounded growth in cumulative PPL beyond the training length, 
    with local spikes exceeding 7000; its evaluation on \texttt{mixed\_chat} encounters numerical overflow (NaN) beyond 20M tokens, so baseline curves are truncated to the finite region, whereas both SANE models remain finite over the entire stream.
    SANE ($\alpha$=5, blue) maintains nearly flat cumulative PPL throughout, 
    while SANE ($\alpha$=10, orange) shows a mild cumulative PPL uptrend at 100M tokens, 
    foreshadowing the collapse observed in Table~\ref{tab:main}. 
    Extreme outliers are clipped at 4$\sigma$ for visualization clarity.}
    \label{fig:ppl-extrapolation}
\end{figure*}

\begin{figure*}[t]
  \centering
  \includegraphics[width=\textwidth]{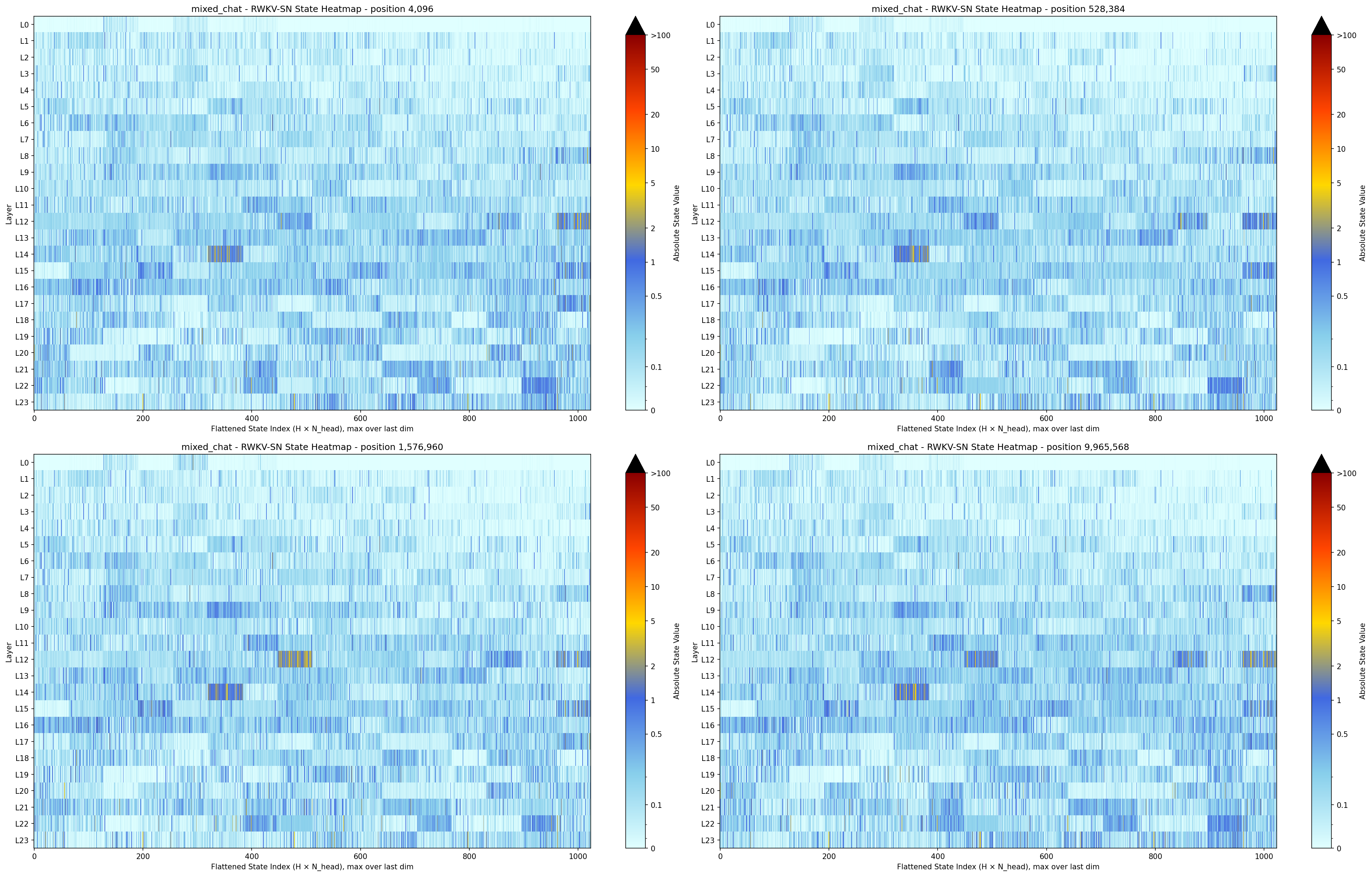}
  \caption{State sparsity preservation in RWKV-SANE ($\alpha$=5) across increasing context lengths. 
  Heatmaps show absolute state absolute values (flattened across heads and channels) for all 24 layers. 
  \textbf{Top-left:}4,096 tokens. 
  \textbf{Top-right:} 528K tokens. 
  \textbf{Bottom-left:} 1.5M tokens. 
  \textbf{Bottom-right:} 9.9M tokens. 
  Crucially, even at $\sim$10M tokens, the state remains highly sparse with values 
  concentrated below 1.0 (light blue), and no head exhibits the uncontrolled 
  red/black hotspots seen in the baseline (Figure~\ref{fig:state-evolution}).}
  \label{fig:sn-state-health}
\end{figure*}

\begin{table*}[t]
\centering
\small
\setlength{\tabcolsep}{3pt}
\caption{Mathematical reasoning performance under short-context (no prefix, Ctx=0) and ultra-long-context (100M prefix) settings. 
At 100M tokens, the baseline collapses entirely. 
Paired t-test p-values (relative to the RWKV-0.4B baseline at Ctx=0, across the 11 benchmarks) are reported as subscripts on the model names.}
\label{tab:main}
\begin{tabular}{@{}lccccccccccccc@{}}
\toprule
\textbf{Model} & \textbf{Ctx} & \textbf{ASDiv} & \textbf{Carpe} & \textbf{CMATH} & \textbf{Col.M} & \textbf{GaoKao} & \textbf{GSM8k} & \textbf{M500} & \textbf{MAWPS} & \textbf{Min.} & \textbf{SVAMP} & \textbf{Weak} & \textbf{Avg} \\
\midrule
RWKV-0.4B & 0 & 78.87 & 35.66 & 60.33 & 16.04 & 31.05 & 57.16 & 42.0 & 89.35 & 4.04 & 39.4 & 28.8 & 43.88 \\
\midrule
\multirow{2}{*}{+SANE ($\alpha$=1)$_{\text{\tiny p<0.001}}$} 
 & 0 & 75.76 & 32.27 & 56.83 & 15.22 & 31.62 & 54.51 & 37.4 & 87.22 & 3.68 & 36.1 & 26.2 & 41.53 \\
 & 100M & 69.89 & 31.56 & 54.33 & 13.98 & 29.91 & 46.25 & 32.4 & 84.46 & 2.21 & 25.7 & 20.7 & 37.40 \\
\midrule
\multirow{2}{*}{+SANE ($\alpha$=2)$_{\text{\tiny p=0.006}}$} 
 & 0 & 76.34 & 33.81 & 57.83 & 15.33 & 32.76 & 55.27 & 40.8 & 88.28 & 3.31 & 38.0 & 27.8 & 42.68 \\
 & 100M & 68.08 & 31.45 & 48.0 & 13.63 & 30.2 & 42.15 & 29.2 & 84.41 & 2.57 & 27.5 & 16.3 & 35.77 \\
\midrule
\multirow{2}{*}{+SANE ($\alpha$=3)$_{\text{\tiny p=0.539}}$} 
 & 0 & 77.83 & 34.73 & 58.33 & 15.58 & 34.76 & 56.71 & 40.2 & 88.91 & 4.78 & 38.5 & 29.1 & 43.58 \\
 & 100M & 68.31 & 29.41 & 48.67 & 13.84 & 27.64 & 42.23 & 33.8 & 83.34 & 2.94 & 22.9 & 18.1 & 35.56 \\
\midrule
\multirow{2}{*}{+SANE ($\alpha$=4)$_{\text{\tiny p=0.754}}$} 
 & 0 & 77.34 & 33.71 & 60.17 & 16.22 & 32.76 & 55.34 & 41.6 & 88.57 & 4.78 & 41.2 & 29.6 & 43.75 \\
 & 100M & 65.91 & 30.43 & 48.0 & 14.27 & 27.07 & 37.07 & 28.2 & 83.97 & 2.57 & 28.4 & 16.0 & 34.72 \\
\midrule
\multirow{2}{*}{+SANE ($\alpha$=5)$_{\text{\tiny p=0.154}}$} 
 & 0 & 77.97 & 33.91 & 60.17 & 15.37 & 32.19 & 56.41 & 40.6 & 89.49 & 4.41 & 38.1 & 29.4 & 43.46 \\
 & 100M & 65.37 & 28.18 & 46.83 & 12.88 & 25.93 & 34.8 & 25.2 & 82.81 & 2.21 & 27.3 & 16.6 & 33.46 \\
\midrule
\multirow{2}{*}{+SANE ($\alpha$=6)$_{\text{\tiny p=0.202}}$} 
 & 0 & 76.57 & 34.94 & 57.33 & 15.65 & 32.48 & 56.03 & 43.8 & 89.01 & 4.04 & 38.2 & 28.3 & 43.30 \\
 & 100M & 50.34 & 22.13 & 34.5 & 10.93 & 27.92 & 18.27 & 20.4 & 63.34 & 2.94 & 11.2 & 7.3 & 24.48 \\
\midrule
\multirow{2}{*}{+SANE ($\alpha$=7)$_{\text{\tiny p=0.994}}$} 
 & 0 & 77.74 & 34.22 & 59.50 & 15.72 & 32.19 & 58.15 & 44.0 & 87.80 & 4.41 & 39.2 & 29.8 & 43.88 \\
 & 100M & 40.32 & 15.37 & 22.33 & 6.85 & 27.07 & 10.54 & 9.0 & 50.17 & 1.10 & 14.8 & 5.4 & 18.45 \\
\midrule
\multirow{2}{*}{+SANE ($\alpha$=8)$_{\text{\tiny p=0.581}}$} 
 & 0 & 76.66 & 33.61 & 60.50 & 15.37 & 32.19 & 59.59 & 43.6 & 88.62 & 5.15 & 39.8 & 30.5 & 44.14 \\
 & 100M & 0.05 & 0.2 & 0 & 0.07 & 1.14 & 0.08 & 0 & 0 & 0 & 0.1 & 0 & 0.15 \\
\midrule
\multirow{2}{*}{+SANE ($\alpha$=9)$_{\text{\tiny p=0.707}}$} 
 & 0 & 77.52 & 34.22 & 59.33 & 15.58 & 37.89 & 58.30 & 42.0 & 89.06 & 5.51 & 39.6 & 26.8 & 44.16 \\
 & 100M & 0 & 0.1 & 0 & 0 & 0 & 0 & 0.2 & 0 & 0 & 0 & 0 & 0.03 \\
\midrule
\multirow{2}{*}{+SANE ($\alpha$=10)$_{\text{\tiny p=0.250}}$} 
 & 0 & 78.56 & 35.04 & 60.50 & 16.04 & 31.91 & 58.15 & 43.0 & 88.18 & 4.41 & 39.6 & 31.1 & 44.23 \\
 & 100M & 0 & 0 & 0 & 0 & 0.28 & 0 & 0 & 0 & 0 & 0 & 0 & 0.03 \\
\bottomrule
\end{tabular}
\end{table*}

\begin{figure*}[t]
  \centering
  \includegraphics[width=\textwidth]{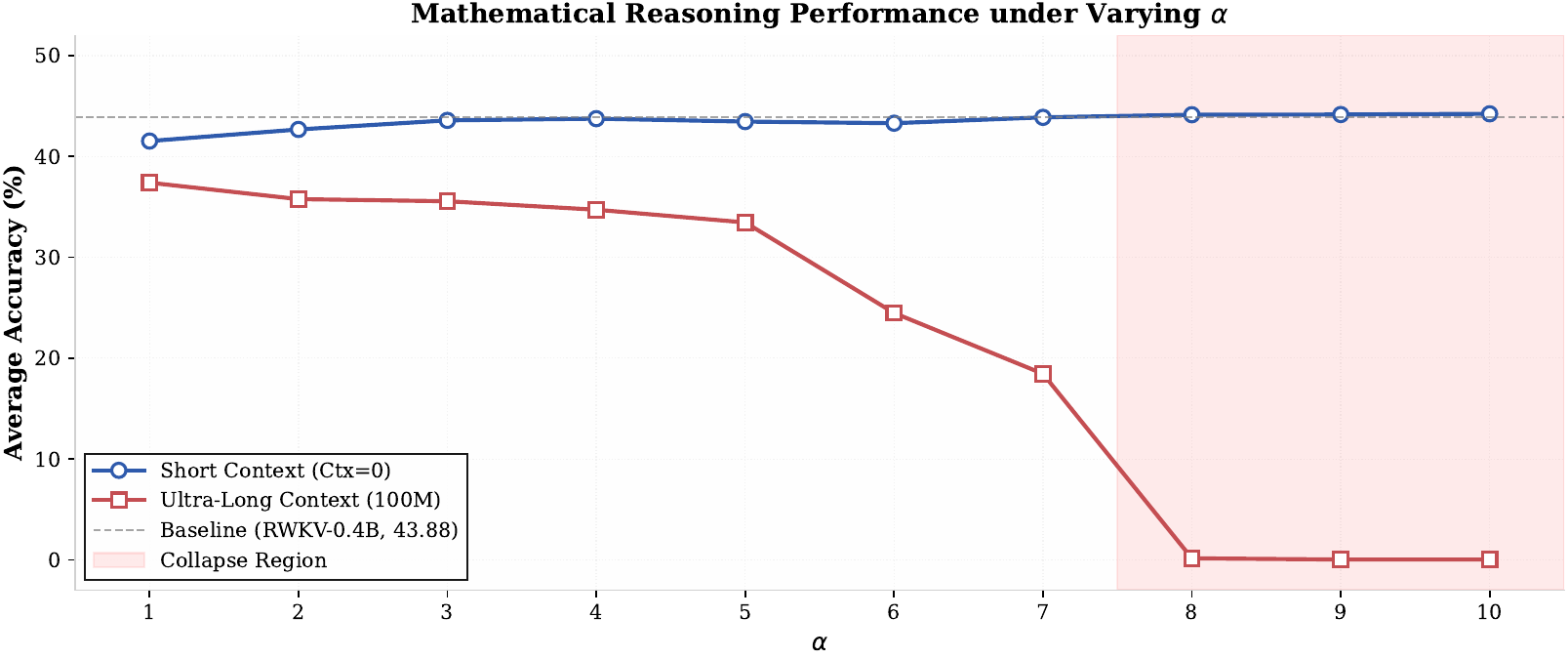}
  \caption{Average accuracy vs.\ threshold scale $\alpha$. Blue: Ctx=0; red: 100M; dashed gray: baseline ($43.88$). Shaded region: collapse regime ($\alpha{\ge}8$).}
  \label{fig:scaling-analysis}
\end{figure*}

\section{Experiments}
\label{sec:experiments}

\subsection{Experimental Setup}
\label{sec:setup}

All experiments in this paper are conducted on RWKV-7, which, as discussed in Section~\ref{sec:preliminary}, is a concrete instantiation of the affine Delta-Rule update. Among purely recurrent Delta-Rule models, RWKV-7 offers a production-grade 0.4B pretrained checkpoint with mature open-source tooling, which enables controlled fine-tuning experiments under a limited computational budget (four NVIDIA RTX 4090 GPUs). We therefore use it as the experimental testbed for validating SANE.

\textbf{Evaluation strategy: why mathematical reasoning.}
It is important to clarify our evaluation philosophy. Delta-Rule models use a fixed-size state matrix to store historical information; this physical constraint implies that forgetting is inevitable. Therefore, the goal of evaluating ultra-long contexts should not be to pursue long-text memorization, but rather to examine whether the model can maintain stable reasoning capabilities after processing an extremely long prefix. Mathematical reasoning tasks naturally align with this philosophy: they do not require verbatim recall of specific facts from the prefix, but instead demand that the model still performs precise symbolic computation after the state has undergone an ultra-long evolution. Grounded in this stance, we adopt mathematical reasoning as the core evaluation probe throughout the paper.

\textbf{Evaluation data.}
We employ the following 11 public mathematical evaluation benchmarks: GSM8k~\cite{cobbe2021training}, Weak12k~\cite{liang2023generalizing}, MATH-500~\cite{lightman2024let}, ASDiv~\cite{miao2020diverse}, Carpe-en~\cite{zhang2023evaluating}, CMATH~\cite{wei2023cmath}, CollegeMath~\cite{tang2024mathscale}, GAOKAO Math QA (abbreviated as GaoKao)~\cite{zhong2024agieval}, MAWPS~\cite{koncel2016mawps}, Minerva~\cite{lewkowycz2022solving}, and SVAMP~\cite{patel-etal-2021-nlp}.
These datasets cover both Chinese and English, include multiple-choice and open-ended questions, and span difficulty levels from elementary school to university. They have been adopted by numerous large-model reasoning studies~\cite{lin2026cmcts,yang2024qwen2,shao2024deepseekmath,guanrstar,qi2025mutual,lin2024large}, enabling a comprehensive assessment of model reasoning capabilities. To save space, we use the following abbreviations in the tables: MATH-500 as M500, Carpe-en as Carpe, CollegeMath as Col.M, Minerva as Min., and Weak12k as Weak.

\textbf{Training data.}
We construct approximately 2.37M supervised fine-tuning (SFT) instances with sequence lengths not exceeding 4097 (the training length is 4096, and one additional token accommodates the one-token shift in autoregressive training), curated from AM-Thinking-v1-Distilled~\cite{tian2025correctanswersequaldistillation}, OpenThoughts3-1.2M~\cite{guha2025openthoughts}, Chinese-DeepSeek-R1-Distill-data-110k~\cite{Chinese-Data-Distill-From-R1}, OpenMathReasoning~\cite{moshkov2025aimo}, DeepMath-103K~\cite{he2025deepmath}, and OpenR1-Math-220k~\cite{openr1}.

\textbf{Ultra-long context synthetic data.}
To examine extrapolation stability at extreme lengths, we concatenate the training split of DeepMath-103K to synthesize a single long text of approximately 10M tokens, denoted as \texttt{english\_math\_chat}. Furthermore, we randomly mix and concatenate DeepMath-103K and Chinese-DeepSeek-R1-Distill-data-110k to construct a mixed long text of approximately 100M tokens, denoted as \texttt{mixed\_chat}.

\textbf{Model configurations.}
Under strictly controlled variables, we initialize from the same pretrained checkpoint (RWKV-7-0.4B-G1D) and fine-tune eleven configurations: the baseline (RWKV-7-0.4B) and ten SANE-augmented variants (RWKV-SANE-0.4B) with $\alpha \in \{1,2,\dots,10\}$. All models are trained with identical data, hyperparameters, and decoding procedures; the only difference is whether the SANE module is introduced and the value of the threshold scale $\alpha$. The visualizations in Figure~\ref{fig:long-ppl} and Figure~\ref{fig:state-evolution} are both based on \texttt{mixed\_chat} and rendered using the fine-tuned baseline model. Complete training hyperparameters are provided in the appendix.

Given the modest model scale (0.4B), greedy decoding tends to produce repetitive outputs that do not reflect true reasoning capability. Therefore, both short-context and extreme-length reasoning adopt sampling-based decoding (top-p $=0.8$, top-k $=20$, temperature $=1.0$), with majority voting over 8 samples (maj@8) to mitigate stochastic variability.

Based on the above setup, our experiments are organized into three parts:
\begin{enumerate}
    \item Evaluate the base reasoning capability on the 11 mathematical benchmarks without any prefix;
    \item Compute the perplexity of different models on \texttt{english\_math\_chat} and \texttt{mixed\_chat};
    \item First feed the model a prefix of approximately 100M tokens of mixed mathematical long text (\texttt{mixed\_chat}), then evaluate the mathematical benchmarks under the identical protocol as in the short-context setting, to verify whether the model still retains symbolic reasoning capability after processing an ultra-long context.
\end{enumerate}
\subsection{Short-Context Mathematical Reasoning}
\label{sec:short-context}

As shown in the Ctx=0 section of Table~\ref{tab:main}, we report zero-shot results on the 11 mathematical benchmarks. Overall, SANE does not sacrifice short-context capability within a safe threshold regime, while measurable differences remain both outside this regime and across different threshold scales within it.

\textbf{SANE does not sacrifice short-context capability within the safe threshold regime.}
The baseline RWKV-7-0.4B achieves an average accuracy of $43.88$. For $\alpha \ge 3$, the SANE variants attain average accuracies between $43.30$ ($\alpha{=}6$) and $44.23$ ($\alpha{=}10$), and paired $t$-tests indicate that none of these configurations differs from the baseline with statistical significance (all $p \ge 0.154$). This shows that within the range $\alpha \ge 3$, the soft compression at chunk boundaries has no measurable impact on short-context reasoning, and SANE can be deployed without sacrificing short-context capability. However, for $\alpha \le 2$, performance is significantly lower than the baseline: $\alpha{=}2$ attains $42.68$ ($p=0.006$), and $\alpha{=}1$ attains $41.53$ ($p<0.001$), indicating that overly aggressive compression impairs short-context reasoning. Therefore, the safe threshold regime for short-context performance is $\alpha \ge 3$.

\textbf{Threshold scale still produces measurable differences in short-context performance.}
These differences are most directly visible at the low-threshold end: $\alpha{=}1$ and $\alpha{=}2$ are significantly below the baseline, indicating that excessively strong compression disrupts effective features acquired during pretraining. At the same time, a mild scale effect can still be observed within the safe regime: the gap between $\alpha{=}10$ and $\alpha{=}5$ is statistically significant ($p=0.038$), showing that the threshold scale has a substantive effect on short-context capacity. This gap can be explained by the spatial structure of the state heatmaps: comparing the baseline (Figure~\ref{fig:state-evolution}, top-left, within training length) and SANE with $\alpha{=}5$ (Figure~\ref{fig:sn-state-health}, top-left), the high-activation regions of the two models overlap substantially in space (e.g., corresponding positions in layers L14 and L21). These regions correspond to legitimate high-value features acquired during pretraining, rather than extreme outliers; $\alpha{=}5$ mildly compresses them, which manifests as an average gap of $0.42$ points, whereas the higher saturation ceiling of $\alpha{=}10$ preserves these legitimate large values more completely below the compression threshold, rendering it numerically slightly above the baseline and statistically indistinguishable from it.

It is worth noting that the above results are obtained in a plug-and-play setting: we introduce SANE into an already pretrained checkpoint via SFT, rather than pretraining from scratch. Inserting a new module with its own parameters ($W_\tau$) and a nonlinear transformation ($\tanh$) inevitably perturbs the numerical distribution of the pretrained weights; the model requires a certain number of SFT steps to adapt to this intervention. After convergence, short-context performance is almost fully recovered, as evidenced by the statistically insignificant difference from the baseline. The residual numerical gap reflects an inherent adaptation cost of the post-hoc SFT strategy. We hypothesize that if SANE were integrated during pretraining, this adaptation cost might be absorbed by the standard optimization process, but this conjecture remains to be verified experimentally. Therefore, under the assumption that the pretraining-integration hypothesis holds, the current short-context results can be viewed as a conservative reference for SANE's short-context performance.

\subsection{Perplexity Evolution Under Extreme-Length Extrapolation}
\label{sec:ppl-evolution}

\paragraph{Baseline: unbounded norm growth and persistent perplexity degradation.}
On \texttt{english\_math\_chat} ($\sim$10M tokens), the baseline's cumulative perplexity climbs steadily after crossing the training-length boundary, showing no sign of saturation, reaching approximately $175$ at 10M tokens, far above its short-context level.

On \texttt{mixed\_chat} ($\sim$100M tokens), the degradation intensifies further: the evaluation encounters numerical overflow (NaN) at around 20M tokens, so only the finite prefix is displayed; within this finite prefix, the cumulative perplexity already exceeds $580$, with local perplexity spikes beyond $7{,}000$ (Figure~\ref{fig:ppl-extrapolation}, left column), while the state Frobenius norm grows monotonically and surpasses $10^{19}$ (Figure~\ref{fig:long-ppl}). This trajectory is a direct empirical corroboration of the mechanistic analysis in Section~\ref{sec:preliminary}: sustained low-rank injections are repeatedly amplified by the cumulative transition product, and the model itself cannot purge the contamination (as diagnosed in Section~\ref{sec:diagnosis}), so the growth continues until overflow.

\paragraph{SANE: always bounded, but the stabilized level is determined by the threshold scale.}
Once SANE is applied, the picture changes qualitatively. With $\alpha{=}5$, the cumulative perplexity remains nearly flat over the entire 100M-token stream, staying close to its short-context level; at around 10M tokens, the state itself remains relatively sparse, with values concentrated below $1.0$ and no runaway hotspots (Figure~\ref{fig:sn-state-health}). With $\alpha{=}10$, the perplexity is likewise bounded; however, its perplexity level is overall higher than that of $\alpha{=}5$ (Figure~\ref{fig:ppl-extrapolation}, right column). We observe that the perplexity converges once the sequence reaches a certain length. This is the most direct empirical evidence for the boundary-wise magnitude control of Section~\ref{subsec:sane-operation}. For visual clarity, Figure~\ref{fig:ppl-extrapolation} presents only these two representative configurations ($\alpha{=}5$ and $\alpha{=}10$); the complete set of threshold scales is evaluated in Table~\ref{tab:main}.

The two configurations together corroborate the central claim of this paper: the baseline's state norm grows without bound and its perplexity collapses accordingly, whereas SANE remains bounded over the entire 100M-token stream. However, it is important to note that although $\alpha{=}10$ does not explode, it settles at an elevated level that can no longer support symbolic reasoning. This means that while SANE can neutralize the contamination in the state, once the neutralization threshold is set too high, the model still resides in an unhealthy state.

\subsection{Extrapolation Reasoning: Functional Reasoning at 100M Tokens}
\label{sec:100m-reasoning}

Numerical boundedness does not by itself guarantee that the model can still reason. We first feed the model approximately 100M tokens of mixed mathematical long text (\texttt{mixed\_chat}, over $24{,}000\times$ the training length), and then evaluate symbolic reasoning on the 11 mathematical benchmarks under the identical protocol as in the short-context setting (top-$p$ sampling, maj@8). The baseline has already encountered NaN at this length, so evaluating it is no longer meaningful.

Figure~\ref{fig:scaling-analysis} presents the 100M reasoning results across the full threshold spectrum: average accuracy decreases monotonically with increasing $\alpha$, revealing three qualitatively distinct regimes.

In the safe regime ($\alpha \le 5$), all configurations retain functional reasoning, with average accuracy declining gradually from $37.40$ at $\alpha{=}1$ to $33.46$ at $\alpha{=}5$. Among these, $3 \le \alpha \le 5$ constitutes the lossless safe regime: as shown in Section~\ref{sec:short-context}, these configurations exhibit no statistically significant difference from the baseline in short contexts while retaining functional reasoning after 100M tokens. Furthermore, $\alpha{=}3$ is Pareto-optimal within the tested range: it attains the best 100M average accuracy ($35.56$) among all configurations that preserve short-context performance ($p=0.539$).

In the steep-decline regime ($\alpha{=}6$--$7$), the first sharp drop occurs: $\alpha{=}6$ falls to $24.48$, and $\alpha{=}7$ further declines to $18.45$. Relative to its short-context level ($43.88$), $\alpha{=}7$ drops by more than 25 points, yet it has not collapsed to zero, indicating that it still retains symbolic reasoning capability.

In the collapse regime ($\alpha \ge 8$), all configurations score near zero on average ($0.15$ or below), while their perplexity remains bounded over the entire 100M-token stream (Section~\ref{sec:ppl-evolution}): numerically healthy, functionally dead.

The mechanism underlying this three-regime degradation can be attributed to the threshold scale's ability to intercept early anomalies. When the threshold scale is too permissive ($\alpha \ge 8$), anomalous values escape compression at low magnitudes, accumulate across chunks, and eventually push the state into an unhealthy operating range---the perplexity no longer explodes, but the precise numerical relationships required for rigorous symbolic computation are disrupted. $\alpha{=}6$ and $\alpha{=}7$ provide transitional evidence for this mechanism: the extent of anomaly accumulation determines the depth of degradation. The collapse regime here serves as a destructive control relative to the safe regime, demonstrating that numerical stability and functional health are two independent dimensions.

In summary, configurations within the safe regime ($\alpha \le 5$) retain functional symbolic reasoning capability after a context over $24{,}000\times$ the training length, among which $\alpha{=}3$ is Pareto-optimal, endowing the boundary-wise magnitude control with functional meaning. The steep-decline and collapse regimes together reveal a continuous trade-off between capacity and stability: numerical stability does not necessarily imply functional health. This also reflects the essence of SANE---neutralizing the contamination beyond the threshold, rather than eliminating it.

\section{Analysis}
\label{sec:analysis}

The experimental results in the preceding sections demonstrate that SANE, within a safe threshold range, preserves short-context capability while maintaining functional reasoning after 100M-token extrapolation; meanwhile, the choice of threshold scale induces a systematic trade-off between capacity and stability. This section explains these phenomena from a mechanistic perspective. We first analyze the structural sources of relative sparsity, then explain why SANE's scale selectivity is effective, and finally discuss the capacity--stability dilemma in threshold scale selection.

\subsection{Sources of Relative Sparsity}
\label{sec:relative-sparsity-sources}

In our diagnosis (Section~\ref{sec:diagnosis}), we empirically observed that the state matrix exhibits relative sparsity: the magnitudes of most entries remain within a small range, while only a few entries deviate substantially from the background. This section explains the structural sources of this phenomenon from the perspective of the online update mechanism of Delta-Rule models. It should be emphasized that the discussion here is a mechanistic interpretation rather than a rigorous sparsity guarantee.

Delta-Rule models typically treat the state matrix as a parameter updated online token by token. From the SGD perspective, the state matrix receives a gradient signal at each time step and updates itself along that direction. Within this framework, a subset of Delta-Rule models, such as RWKV-7~\cite{peng2023rwkv,peng2025rwkv,peng2024eagle} and GDN~\cite{yang2026beyond,cai2026u}, introduce an additional weight decay term during the update. Weight decay is equivalent to L2 regularization from the SGD perspective: it exerts a persistent shrinkage pressure proportional to the current value on all state entries, driving the overall state toward smaller magnitudes. This shrinkage pressure is particularly effective on directions that are infrequently updated---those directions that rarely participate in computation lack sufficient injection to counteract the shrinkage and are thus pressed into a low-magnitude background. Therefore, L2 shrinkage explains the ``mostly small values'' aspect of relative sparsity.

However, shrinkage pressure alone does not determine where large values may appear. This requires understanding the structural characteristics of the update. For each head, the state matrix $S_t \in \mathbb{R}^{d \times d}$ has $d^2$ entries, but the injection term $K_t$ at each time step is low-rank (rank-1 in RWKV-7, specifically $v_t^\top k_t$). This means that a single-step update can only directly affect a low-dimensional subspace of the state matrix. In other words, the role of the low-rank update is not to guarantee that large values will appear, but to place an upper bound on the spatial range where large values can emerge: if certain directions are to maintain large values under persistent L2 shrinkage, they can only appear in the directions touched by the low-rank update, and cannot spread across all directions of the entire state matrix. Here, ``directions'' refers to the low-dimensional subspace affected by the update, rather than sparsity in matrix element positions; a low-rank matrix itself can be dense, and the number of directions touched can grow after multi-step accumulation. The low-rank update merely restricts the range of updates that can counteract shrinkage and sustain large values at each time step.

Therefore, relative sparsity can be understood as the result of competition between two forces: L2 shrinkage provides a global, persistent pressure toward small magnitudes, pressing most infrequently updated directions into a low-magnitude background; the low-rank update restricts the directions capable of counteracting shrinkage and maintaining large values to a small number of low-dimensional subspaces. Whether large values can ultimately appear and persist in a given direction depends on whether that direction receives sufficiently repeated injections over the sequence---a matter determined jointly by data distribution and training dynamics rather than by the low-rank structure alone. The competition between these two forces stably maintains an order-of-magnitude separation in the state matrix---this is precisely the ``relatively sparse substrate'' diagnosed in Section~\ref{sec:diagnosis}.

\subsection{Scale Selectivity}
\label{sec:scale-selectivity}

The experimental results in Section~\ref{sec:experiments} raise a deeper question: why can SANE effectively suppress state norm explosion under extreme lengths while incurring almost no performance loss in short contexts? To answer this question, we need to return to the sources of relative sparsity analyzed in Section~\ref{sec:relative-sparsity-sources}---the competition between L2 shrinkage and low-rank updates maintains an order-of-magnitude separation in the state matrix. This structural property is precisely the foundation upon which SANE achieves \textbf{scale selectivity}.

Specifically, scale selectivity refers to SANE's ability to distinguish between two numerical scales in the state matrix---small background values and anomalous large values---and to apply compression only to the latter. This capability rests on two jointly necessary conditions.

First, relative sparsity provides the structural prerequisite for selective truncation. As analyzed in Section~\ref{sec:relative-sparsity-sources}, L2 shrinkage continuously presses most infrequently updated directions into a low-magnitude background, while low-rank updates restrict the directions capable of sustaining large values to a small number of low-dimensional subspaces. The result is that in a healthy state matrix, most values lie below $0.1$, and only a few directions may accumulate substantially larger values. This natural order-of-magnitude gap between large values and the background means that a simple magnitude-based threshold suffices to distinguish them, without requiring the model to learn a complex decision boundary. The heatmap analysis in Section~\ref{sec:short-context} provides direct evidence for this: the high-activation regions of the baseline and SANE ($\alpha{=}5$) overlap substantially in space (e.g., corresponding positions in layers L14 and L21), indicating that SANE preserves the locations and sources of legitimate large values while gently compressing extreme values.

Second, the nonlinearity of $\tanh$ converts magnitude differences into selective compression. When $|S|/\tau < 0.1$, $\tanh(S/\tau) \approx S/\tau$, an approximate identity mapping. This holds for all background values because the diagnosis shows that healthy background values concentrate below $0.1$, and $\tau > 1$ guarantees that $|S|/\tau < 0.1$ is always satisfied. When $|S|/\tau \gg 1$, $\tanh$ saturates and the output is softly truncated to $\pm\tau$---only anomalous large values enter this regime. The smooth transition between the two regimes ensures differentiability: gradients can still flow through the saturation regime, allowing the model to learn to adjust the threshold dynamically according to context during training.

From the cumulative product structure of state evolution in Section~\ref{sec:preliminary}, under extreme extrapolation this cumulative product chain produces unidirectional amplification in specific channels. The $\tanh$ compression applied by SANE at chunk boundaries interrupts this cumulative chain---anomalous values are compressed back to a bounded range before they can fully amplify, while the intra-chunk parallel computation structure remains entirely unaffected.

The full threshold spectrum (Figure~\ref{fig:scaling-analysis}) provides macroscopic verification of this mechanism: within the safe regime ($\alpha \le 5$), selective compression is sufficient to suppress anomalous growth while preserving functional reasoning; as $\alpha$ increases further, the compression strength becomes insufficient to cover the magnitude of anomalies, and the mechanism fails. This experimentally confirms that scale selectivity is the core reason for SANE's effectiveness.

In summary, the essence of scale selectivity is this: the relative sparsity analyzed in Section~\ref{sec:relative-sparsity-sources} provides the structural prerequisite for selective compression, the $\tanh$ nonlinearity converts this magnitude separation into selective compression, and the application at chunk boundaries ensures that contamination is neutralized before cumulative amplification can take hold.

\subsection{The Capacity--Stability Dilemma}
\label{sec:dilemma}

The full threshold spectrum in Section~\ref{sec:experiments} reveals a thought-provoking phenomenon: the same threshold scale $\alpha$ plays diametrically opposite roles in short-context and long-context settings. In the short-context setting, higher thresholds ($\alpha \ge 7$) are nearly lossless or even slightly better than the baseline ($\alpha{=}7$ attains $43.88$, $p=0.994$; $\alpha{=}10$ attains $44.23$, $p=0.250$), whereas overly low thresholds ($\alpha \le 2$) significantly impair performance ($\alpha{=}1$ attains $41.53$, $p<0.001$; $\alpha{=}2$ attains $42.68$, $p=0.006$). After 100M-token extrapolation, however, the situation reverses completely: all configurations within the safe regime ($\alpha \le 5$) retain functional reasoning ($33.46$--$37.40$), $\alpha{=}6$--$7$ exhibit a steep decline ($24.48$ and $18.45$), and configurations with $\alpha \ge 8$ lose reasoning capability entirely despite bounded perplexity (average accuracy $\le 0.15$). This systematic reversal between short-context and long-context behavior reveals a fundamental dilemma.

\textbf{Short-context perspective: why higher thresholds perform better.}
$\alpha$ controls the saturation ceiling of the $\tanh$ compression: the larger $\alpha$ is, the higher the threshold $\tau$, and the more completely legitimate large values are preserved. Within the training length, legitimate high-value features acquired during pretraining (with magnitudes around $5$--$20$) carry useful reasoning-relevant information. When $\alpha$ is sufficiently high (e.g., $\alpha{=}7$, $\alpha{=}10$), these legitimate large values fall almost entirely within the linear regime and remain uncompressed, so short-context capability is indistinguishable from the baseline. Conversely, when $\alpha$ is too low (e.g., $\alpha{=}1$, $\alpha{=}2$), legitimate large values are over-compressed, and short-context performance degrades significantly. This explains the sensitivity of short-context performance to the lower bound of the threshold: the threshold must not be too low, or pretrained capability is impaired.

\textbf{Long-context perspective: why high thresholds fail.}
However, the long-context setting introduces a challenge absent in short contexts: anomalous values do not emerge at magnitudes exceeding $100$ from the outset. During the early stages of extrapolation, anomalies grow gradually from the healthy range and must pass through an intermediate stage where their magnitudes overlap with those of legitimate large values (approximately $10$--$50$). High-threshold configurations ($\alpha \ge 8$) cannot distinguish these intermediate-magnitude early anomalies from legitimate large values---both fall within the linear regime and pass through almost losslessly. These early anomalies continue to accumulate as the sequence grows, propagate across chunks, amplify layer by layer, and ultimately push the state into an unhealthy operating range: the perplexity no longer explodes, but the precise numerical relationships required for rigorous symbolic computation have been disrupted, and reasoning capability is completely lost. $\alpha{=}6$ and $\alpha{=}7$ provide transitional evidence for this mechanism: their thresholds are tighter than those of $\alpha \ge 8$, so early anomalies are partially compressed, and consequently the functional degradation is gradual ($24.48$ and $18.45$) rather than complete ($\le 0.15$)---the extent of anomaly accumulation determines the depth of degradation.

\textbf{Low thresholds' survival: the cost and benefit of scale discrimination.}
Low-threshold configurations ($\alpha \le 5$) lower the saturation ceiling, causing anomalous values to enter the transition or saturation regime of $\tanh$ at an earlier stage, where they are effectively truncated and prevented from accumulating across chunks to destructive magnitudes. This lower ceiling also acts on legitimate large values---pretrained features in the $5$--$20$ range experience mild compression, which is precisely why $\alpha{=}5$ is slightly weaker than the baseline in the short-context setting. However, the degree of compression is sufficiently mild that it does not harm overall reasoning capability: the average gap of $0.42$ points is not statistically significant ($p=0.154$). On the long-context side, this cost buys a decisive benefit. The full spectrum further shows that the safe regime is not a single-point phenomenon: $\alpha{=}3$, $\alpha{=}4$, and $\alpha{=}5$ retain average accuracies of $35.56$, $34.72$, and $33.46$, respectively. Among these, $\alpha{=}3$ is Pareto-optimal within the tested range---it attains the best 100M performance among all configurations that preserve short-context capability ($p=0.539$). $\alpha{=}1$ and $\alpha{=}2$, although scoring highest at 100M, significantly sacrifice short-context capability and therefore do not belong to the lossless safe regime.

The above phenomena point to a structural dilemma of SANE. Legitimate large values ($5$--$20$) and early-stage anomalies ($10$--$50$) overlap in numerical range. Although the threshold $\tau$ itself is an input-dependent dynamic quantity, $\alpha$ controls the overall compression strength---it determines the scale ceiling of the dynamic threshold, but cannot alter this strength during inference. Anomalous values grow monotonically with extrapolation distance, yet the compression strength cannot adaptively tighten or relax in response. An excessively high $\alpha$ chooses to preserve capacity at the cost of letting early anomalies pass through in long contexts, ultimately leading to functional collapse; an excessively low $\alpha$ chooses to prioritize anomaly suppression at the cost of mildly compressing legitimate large values or even impairing short-context capability; the safe regime ($3 \le \alpha \le 5$) strikes a balance between the two, maintaining a certain level of long-context reasoning capability without sacrificing short-context performance. Among these, $\alpha{=}3$ is Pareto-optimal.

It is worth noting that this dilemma is to a considerable extent a product of post-hoc introduction. The current SANE is introduced via SFT into an already pretrained checkpoint, compressing state representations that were pretrained without compression constraints. The overlap between the magnitude distribution of legitimate large values acquired during pretraining and that of early-stage anomalies is precisely the root of the trade-off. We hypothesize that if SANE were integrated during pretraining, the model could learn representations under the compression constraint, and the magnitude distribution of legitimate features might shift downward overall, thereby narrowing the overlap with early-stage anomalies and alleviating or even eliminating this trade-off. Whether this hypothesis holds remains to be verified by pretraining-from-scratch experiments.

\section{Related Work}
\label{sec:related}

In recent years, linear recurrent neural networks, with RWKV as a prominent representative, have attracted considerable attention~\cite{yang2024gated,huang2026mdn}. These models replace the Transformer's KV cache with a fixed-size recurrent state: delta-rule-based models---including GDN~\cite{yang2025gated,cao2026qwen3,hatamizadeh2026gated}, KDA~\cite{team2025kimi,team2026kimi}, and the RWKV series built upon the Generalized Delta Rule~\cite{peng2023rwkv,peng2025rwkv,peng2024eagle}---together with the Mamba family of state space models~\cite{qu2024survey,yang2025gated,gu2023mamba,lahotimamba} and xLSTM~\cite{beck2024xlstm,auer2026tirex,podest2026tirex}, have demonstrated performance comparable to Transformers on standard tasks. Owing to the absence of the out-of-distribution generalization problem of positional encodings that plagues Transformers, and equipped with built-in temporal decay structures, these models generally maintain stable extrapolation within a few multiples of the training length~\cite{peng2025rwkv}. For Delta-Rule models, however, when the sequence length reaches hundreds or even thousands of times the training length, the numerical instability of the fixed-size state leads to performance collapse. For such extreme-context scenarios, there has been neither a systematic diagnosis of the failure mechanism nor an established solution.

Current research on long-context extrapolation remains predominantly centered on Transformers. One line of work seeks to mitigate the out-of-distribution issues induced by positional encoding extrapolation, including various interpolation and extrapolation strategies for rotary positional embeddings~\cite{aguilar2026does,chen2025hope,shang2025longrope2,huo2026periodic,rahman2025context}; another line directly modifies the attention mechanism itself to enlarge the context window~\cite{liu2025reattention,munkhdalai2024leave}. However, these methods are all built upon the Transformer attention framework, addressing positional encoding problems that Delta-Rule models do not possess. In Delta-Rule models, the sole medium for cross-context information transfer is the fixed-size state matrix, and the bottleneck originates precisely from the state itself. Consequently, these methods cannot be directly transferred.

Existing improvements for RWKV and similar linear RNNs are also difficult to apply directly to ultra-long extrapolation. Current approaches either adopt hybrid architectures with Transformers that require pretraining from scratch~\cite{ren2025samba,behrouz2026titans}, or rely on the selective scan mechanism specific to Mamba-like models for state management~\cite{ben2025decimamba}. These methods either alter the native architecture of Delta-Rule models or depend on components that such models do not possess, and thus cannot achieve extreme-context extrapolation while preserving the standard training and inference pipeline. Therefore, how to endow Delta-Rule models with stable extrapolation to context lengths several orders of magnitude beyond the training length, while maintaining the native architecture and standard pipeline and without sacrificing short-context capability, remains a largely unexplored problem. The State Anomaly Neutralization (SANE) proposed in this paper targets precisely this gap.

\section{Conclusion} 
We investigate the extreme-context failure of RWKV-7 and empirically identify a distinct state pathology: localized norm explosion atop a relatively sparse substrate, rather than global state saturation. Analysis of its recurrent dynamics suggests that uneven injections and transition-induced amplification can dominate state decay in a small number of directions. Motivated by this diagnosis, we propose State Anomaly Neutralization (SANE), which applies adaptive $\tanh$ compression at chunk boundaries to approximately preserve the low-magnitude background while suppressing extreme state values. Introduced into a pretrained RWKV-7 model through supervised fine-tuning, SANE causes no statistically significant degradation on short-context reasoning tasks within a safe threshold range and retains functional symbolic reasoning after a 100M-token prefix, whereas the baseline encounters numerical overflow. The contrast across the full threshold spectrum further reveals a capacity--stability trade-off: preventing numerical explosion alone does not guarantee functional health. These results establish SANE as a lightweight approach to extreme-context state stabilization and motivate its evaluation on broader Delta-Rule architectures.

\section*{Limitations}
\label{sec:limitations}

The empirical validation in this paper is currently confined to RWKV-7 (0.4B parameters), a representative model of the Delta-Rule subset that adopts SGD with L2 regularization. For other models within this subset, SANE is expected to exhibit similar effectiveness because the theoretical guarantee of a sparse substrate is a shared property of the subset; however, this remains to be empirically verified. For Delta-Rule models outside the SGD+L2 subset (e.g., the original Delta-Net), a sparse substrate is not theoretically guaranteed, and the applicability of SANE depends on whether such models exhibit the characteristic pattern of localized norm explosion atop a sparse substrate in practice---a generalization that likewise awaits validation. Extending SANE to a broader range of Delta-Rule architectures and to larger parameter scales (e.g., 7B, 14B) represents an important direction for future work.

Furthermore, this paper introduces SANE into an already pretrained checkpoint via supervised fine-tuning, a plug-and-play setup that inevitably incurs an adaptation cost. As discussed in Section~\ref{sec:short-context}, the current short-context results should be interpreted as a conservative lower bound on SANE's performance. If SANE were integrated during pretraining, this adaptation cost would be absorbed into the standard optimization process, and short-context performance could be further improved. We leave the evaluation of SANE integrated from scratch to future work.

\bibliography{custom}

@inproceedings{peng2023rwkv,
  title={Rwkv: Reinventing rnns for the transformer era},
  author={Peng, Bo and Alcaide, Eric and Anthony, Quentin and Albalak, Alon and Arcadinho, Samuel and Biderman, Stella and Cao, Huanqi and Cheng, Xin and Chung, Michael and Derczynski, Leon and others},
  booktitle={Findings of the association for computational linguistics: EMNLP 2023},
  pages={14048--14077},
  year={2023}
}

@inproceedings{rahman2025context,
  title={Context-Aware Network Resource Allocation for Industry 5.0 Applications},
  author={Rahman, Md Mashiur and Jin, Jiong and Palanisamy, Suresh and van Winckel, Steve and Jayaraman, Prem Prakash and Sardar, Asif Ahmed},
  booktitle={2025 IEEE 35th International Telecommunication Networks and Applications Conference (ITNAC)},
  pages={1--4},
  year={2025},
  organization={IEEE}
}

@article{munkhdalai2024leave,
  title={Leave no context behind: Efficient infinite context transformers with infini-attention},
  author={Munkhdalai, Tsendsuren and Faruqui, Manaal and Gopal, Siddharth},
  journal={arXiv preprint arXiv:2404.07143},
  volume={101},
  pages={15},
  year={2024}
}

@inproceedings{liu2025reattention,
  title={Reattention: Training-free infinite context with finite attention scope},
  author={Liu, Xiaoran and Li, Ruixiao and Liu, Zhigeng and Guo, Qipeng and Song, Yuerong and Lv, Kai and Yan, Hang and Li, Linlin and Liu, Qun and Qiu, Xipeng},
  booktitle={International Conference on Learning Representations},
  volume={2025},
  pages={95458--95478},
  year={2025}
}

@inproceedings{schlag2021linear,
  title={Linear transformers are secretly fast weight programmers},
  author={Schlag, Imanol and Irie, Kazuki and Schmidhuber, J{\"u}rgen},
  booktitle={International conference on machine learning},
  pages={9355--9366},
  year={2021},
  organization={PMLR}
}

@inproceedings{yang2024gated,
  title={Gated Linear Attention Transformers with Hardware-Efficient Training},
  author={Yang, Songlin and Wang, Bailin and Shen, Yikang and Panda, Rameswar and Kim, Yoon},
  booktitle={International Conference on Machine Learning},
  year={2024}
}

@article{behrouz2026titans,
  title={Titans: Learning to memorize at test time},
  author={Behrouz, Ali and Zhong, Peilin and Mirrokni, Vahab},
  journal={Advances in Neural Information Processing Systems},
  volume={38},
  pages={113506--113543},
  year={2026}
}

@inproceedings{ren2025samba,
  title={Samba: Simple hybrid state space models for efficient unlimited context language modeling},
  author={Ren, Liliang and Liu, Yang and Lu, Yadong and Liang, Chen and Chen, Weizhu and others},
  booktitle={International Conference on Learning Representations},
  volume={2025},
  pages={53551--53575},
  year={2025}
}

@inproceedings{chen2025hope,
  title={Hope: A novel positional encoding without long-term decay for enhanced context awareness and extrapolation},
  author={Chen, Yuhan and Lv, Ang and Luan, Jian and Wang, Bin and Liu, Wei},
  booktitle={Proceedings of the 63rd Annual Meeting of the Association for Computational Linguistics (Volume 1: Long Papers)},
  pages={23044--23056},
  year={2025}
}

@article{cobbe2021training,
  title={Training verifiers to solve math word problems},
  author={Cobbe, Karl and Kosaraju, Vineet and Bavarian, Mohammad and Chen, Mark and Jun, Heewoo and Kaiser, Lukasz and Plappert, Matthias and Tworek, Jerry and Hilton, Jacob and Nakano, Reiichiro and others},
  journal={arXiv preprint arXiv:2110.14168},
  year={2021}
}

@inproceedings{patel-etal-2021-nlp,
  title = "Are {NLP} Models really able to Solve Simple Math Word Problems?",
  author = "Patel, Arkil  and
    Bhattamishra, Satwik  and
    Goyal, Navin",
  booktitle = "Proceedings of the 2021 Conference of the North American Chapter of the Association for Computational Linguistics: Human Language Technologies",
  month = jun,
  year = "2021",
  address = "Online",
  publisher = "Association for Computational Linguistics",
  url = "https://aclanthology.org/2021.naacl-main.168",
  doi = "10.18653/v1/2021.naacl-main.168",
  pages = "2080--2094",
}

@article{guha2025openthoughts,
  title={Openthoughts: Data recipes for reasoning models},
  author={Guha, Etash and Marten, Ryan and Keh, Sedrick and Raoof, Negin and Smyrnis, Georgios and Bansal, Hritik and Nezhurina, Marianna and Mercat, Jean and Vu, Trung and Sprague, Zayne and others},
  journal={arXiv preprint arXiv:2506.04178},
  year={2025}
}

@misc{openr1,
    title = {Open R1: A fully open reproduction of DeepSeek-R1},
    url = {https://github.com/huggingface/open-r1},
    author = {{Hugging Face}},
    month = {January},
    year = {2025}
}

@inproceedings{guanrstar,
  title={rStar-Math: Small LLMs Can Master Math Reasoning with Self-Evolved Deep Thinking},
  author={Guan, Xinyu and Zhang, Li Lyna and Liu, Yifei and Shang, Ning and Sun, Youran and Zhu, Yi and Yang, Fan and Yang, Mao},
  booktitle={International Conference on Machine Learning},
  pages={20640--20661},
  year={2025},
  organization={PMLR}
}

@inproceedings{lin2024large,
  title={From large to tiny: Distilling and refining mathematical expertise for math word problems with weakly supervision},
  author={Lin, Qingwen and Xu, Boyan and Huang, Zhengting and Cai, Ruichu},
  booktitle={International Conference on Intelligent Computing},
  pages={251--262},
  year={2024},
  organization={Springer}
}

@inproceedings{qi2025mutual,
  title={Mutual reasoning makes smaller LLMs stronger problem-solver},
  author={Qi, Zhenting and Ma, Mingyuan and Xu, Jiahang and Zhang, Li Lyna and Yang, Fan and Yang, Mao},
  booktitle={International Conference on Learning Representations},
  volume={2025},
  pages={20788--20807},
  year={2025}
}

@article{shao2024deepseekmath,
  title={Deepseekmath: Pushing the limits of mathematical reasoning in open language models},
  author={Shao, Zhihong and Wang, Peiyi and Zhu, Qihao and Xu, Runxin and Song, Junxiao and Bi, Xiao and Zhang, Haowei and Zhang, Mingchuan and Li, YK and Wu, Yang and others},
  journal={arXiv preprint arXiv:2402.03300},
  year={2024}
}

@article{yang2024qwen2,
  title={Qwen2. 5-math technical report: Toward mathematical expert model via self-improvement},
  author={Yang, An and Zhang, Beichen and Hui, Binyuan and Gao, Bofei and Yu, Bowen and Li, Chengpeng and Liu, Dayiheng and Tu, Jianhong and Zhou, Jingren and Lin, Junyang and others},
  journal={arXiv preprint arXiv:2409.12122},
  year={2024}
}

@article{lin2026cmcts,
  title={CMCTS: A Constrained Monte Carlo Tree Search framework for mathematical reasoning in large language model},
  author={Lin, Qingwen and Xu, Boyan and Hu, Guimin and Li, Zijian and Hao, Zhifeng and Zhang, Keli and Cai, Ruichu},
  journal={Applied Intelligence},
  volume={56},
  number={1},
  pages={13},
  year={2026},
  publisher={Springer}
}

@article{he2025deepmath,
  title={Deepmath-103k: A large-scale, challenging, decontaminated, and verifiable mathematical dataset for advancing reasoning},
  author={He, Zhiwei and Liang, Tian and Xu, Jiahao and Liu, Qiuzhi and Chen, Xingyu and Wang, Yue and Song, Linfeng and Yu, Dian and Liang, Zhenwen and Wang, Wenxuan and others},
  journal={arXiv preprint arXiv:2504.11456},
  year={2025}
}

@article{moshkov2025aimo,
  title={Aimo-2 winning solution: Building state-of-the-art mathematical reasoning models with openmathreasoning dataset},
  author={Moshkov, Ivan and Hanley, Darragh and Sorokin, Ivan and Toshniwal, Shubham and Henkel, Christof and Schifferer, Benedikt and Du, Wei and Gitman, Igor},
  journal={arXiv preprint arXiv:2504.16891},
  year={2025}
}

@misc{Chinese-Data-Distill-From-R1,
  author = {Cong Liu and Zhong Wang and ShengYu Shen and Jialiang Peng and Xiaoli Zhang and ZhenDong Du and YaFang Wang},
  title = {The Chinese dataset distilled from DeepSeek-R1-671b},
  year = {2025},
  howpublished = {\url{https://huggingface.co/datasets/Congliu/Chinese-DeepSeek-R1-Distill-data-110k}},
}

@misc{tian2025correctanswersequaldistillation,
      title={Not All Correct Answers Are Equal: Why Your Distillation Source Matters}, 
      author={Xiaoyu Tian and Yunjie Ji and Haotian Wang and Shuaiting Chen and Sitong Zhao and Yiping Peng and Han Zhao and Xiangang Li},
      year={2025},
      eprint={2505.14464},
      archivePrefix={arXiv},
      primaryClass={cs.CL},
      url={https://arxiv.org/abs/2505.14464}, 
}

@article{lewkowycz2022solving,
  title={Solving quantitative reasoning problems with language models},
  author={Lewkowycz, Aitor and Andreassen, Anders and Dohan, David and Dyer, Ethan and Michalewski, Henryk and Ramasesh, Vinay and Slone, Ambrose and Anil, Cem and Schlag, Imanol and Gutman-Solo, Theo and others},
  journal={Advances in neural information processing systems},
  volume={35},
  pages={3843--3857},
  year={2022}
}

@inproceedings{koncel2016mawps,
  title={MAWPS: A math word problem repository},
  author={Koncel-Kedziorski, Rik and Roy, Subhro and Amini, Aida and Kushman, Nate and Hajishirzi, Hannaneh},
  booktitle={Proceedings of the 2016 conference of the north american chapter of the association for computational linguistics: human language technologies},
  pages={1152--1157},
  year={2016}
}

@article{liu2025muon,
  title={Muon is scalable for llm training},
  author={Liu, Jingyuan and Su, Jianlin and Yao, Xingcheng and Jiang, Zhejun and Lai, Guokun and Du, Yulun and Qin, Yidao and Xu, Weixin and Lu, Enzhe and Yan, Junjie and others},
  journal={arXiv preprint arXiv:2502.16982},
  year={2025}
}

@inproceedings{zhong2024agieval,
  title={Agieval: A human-centric benchmark for evaluating foundation models},
  author={Zhong, Wanjun and Cui, Ruixiang and Guo, Yiduo and Liang, Yaobo and Lu, Shuai and Wang, Yanlin and Saied, Amin and Chen, Weizhu and Duan, Nan},
  booktitle={Findings of the association for computational linguistics: NAACL 2024},
  pages={2299--2314},
  year={2024}
}

@article{tang2024mathscale,
  title={Mathscale: Scaling instruction tuning for mathematical reasoning},
  author={Tang, Zhengyang and Zhang, Xingxing and Wang, Benyou and Wei, Furu},
  journal={arXiv preprint arXiv:2403.02884},
  year={2024}
}

@article{wei2023cmath,
  title={CMATH: Can your language model pass Chinese elementary school math test?},
  author={Wei, Tianwen and Luan, Jian and Liu, Wei and Dong, Shuang and Wang, Bin},
  journal={arXiv preprint arXiv:2306.16636},
  year={2023}
}

@article{zhang2023evaluating,
  title={Evaluating and improving tool-augmented computation-intensive math reasoning},
  author={Zhang, Beichen and Zhou, Kun and Wei, Xilin and Zhao, Xin and Sha, Jing and Wang, Shijin and Wen, Ji-Rong},
  journal={Advances in Neural Information Processing Systems},
  volume={36},
  pages={23570--23589},
  year={2023}
}

@inproceedings{miao2020diverse,
  title={A diverse corpus for evaluating and developing English math word problem solvers},
  author={Miao, Shen-Yun and Liang, Chao-Chun and Su, Keh-Yih},
  booktitle={Proceedings of the 58th annual meeting of the Association for Computational Linguistics},
  pages={975--984},
  year={2020}
}

@inproceedings{lightman2024let,
  title={Let's verify step by step},
  author={Lightman, Hunter and Kosaraju, Vineet and Burda, Yuri and Edwards, Harrison and Baker, Bowen and Lee, Teddy and Leike, Jan and Schulman, John and Sutskever, Ilya and Cobbe, Karl},
  booktitle={International Conference on Learning Representations},
  volume={2024},
  pages={39578--39601},
  year={2024}
}

@inproceedings{liang2023generalizing,
  title={Generalizing math word problem solvers via solution diversification},
  author={Liang, Zhenwen and Zhang, Jipeng and Wang, Lei and Wang, Yan and Shao, Jie and Zhang, Xiangliang},
  booktitle={Proceedings of the AAAI Conference on Artificial Intelligence},
  volume={37},
  pages={13183--13191},
  year={2023}
}

@inproceedings{shang2025longrope2,
  title={LongRoPE2: Near-Lossless LLM Context Window Scaling},
  author={Shang, Ning and Zhang, Li Lyna and Wang, Siyuan and Zhang, Gaokai and Lopez, Gilsinia and Yang, Fan and Chen, Weizhu and Yang, Mao},
  booktitle={International Conference on Machine Learning},
  pages={54203--54218},
  year={2025},
  organization={PMLR}
}

@article{aguilar2026does,
  title={Does Your Neural Network Extrapolate? Feature Engineering as Identifiability Bias for OOD Generalization},
  author={Aguilar, Leonel and Nagler, Jan and Hoelscher, Christoph and Antulov-Fantulin, Nino},
  journal={arXiv preprint arXiv:2605.07483},
  year={2026}
}

@article{zhang2026structural,
  title={Structural Calibration of Dual Stream Collaborative Cues Within a Lightweight RWKV Network for Pixel-Level Crack Segmentation},
  author={Zhang, Hanxu and Zheng, Jiangpeng and Jia, Chen and Liu, Hui and Shi, Fan and Cheng, Xu},
  journal={IEEE Transactions on Industrial Informatics},
  year={2026},
  publisher={IEEE}
}

@article{cai2026u,
  title={U-RWKV: Accurate and Efficient Volumetric Medical Image Segmentation via RWKV},
  author={Cai, Hongyu and Wang, Yifan and Wang, Liu and Zhao, Jian and Kuang, Zhejun},
  journal={IEEE Transactions on Image Processing},
  year={2026},
  publisher={IEEE}
}

@inproceedings{yang2026beyond,
  title={Beyond Quadratic: Linear-Time Change Detection with RWKV},
  author={Yang, Zhenyu and Pei, Gensheng and Chen, Tao and Yuan, Xia and Zhang, Haofeng and Shu, Xiangbo and Yao, Yazhou},
  booktitle={Proceedings of the AAAI Conference on Artificial Intelligence},
  volume={40},
  pages={11811--11819},
  year={2026}
}

@article{peng2024eagle,
  title={Eagle and finch: Rwkv with matrix-valued states and dynamic recurrence},
  author={Peng, Bo and Goldstein, Daniel and Anthony, Quentin and Albalak, Alon and Alcaide, Eric and Biderman, Stella and Cheah, Eugene and Du, Xingjian and Ferdinan, Teddy and Hou, Haowen and others},
  journal={arXiv preprint arXiv:2404.05892},
  year={2024}
}

@article{peng2025rwkv,
  title={Rwkv-7" goose" with expressive dynamic state evolution},
  author={Peng, Bo and Zhang, Ruichong and Goldstein, Daniel and Alcaide, Eric and Du, Xingjian and Hou, Haowen and Lin, Jiaju and Liu, Jiaxing and Lu, Janna and Merrill, William and others},
  journal={arXiv preprint arXiv:2503.14456},
  year={2025}
}

@article{beck2024xlstm,
  title={xlstm: Extended long short-term memory},
  author={Beck, Maximilian and P{\"o}ppel, Korbinian and Spanring, Markus and Auer, Andreas and Prudnikova, Oleksandra and Kopp, Michael and Klambauer, G{\"u}nter and Brandstetter, Johannes and Hochreiter, Sepp},
  journal={Advances in Neural Information Processing Systems},
  volume={37},
  pages={107547--107603},
  year={2024}
}

@article{team2025kimi,
  title={Kimi linear: An expressive, efficient attention architecture},
  author={Team, Kimi and Zhang, Yu and Lin, Zongyu and Yao, Xingcheng and Hu, Jiaxi and Meng, Fanqing and Liu, Chengyin and Men, Xin and Yang, Songlin and Li, Zhiyuan and others},
  journal={arXiv preprint arXiv:2510.26692},
  year={2025}
}

@article{qu2024survey,
  title={A survey of mamba},
  author={Qu, Haohao and Ning, Liangbo and An, Rui and Fan, Wenqi and Derr, Tyler and Liu, Hui and Xu, Xin and Li, Qing},
  journal={ACM Transactions on Intelligent Systems and Technology},
  year={2024},
  publisher={ACM New York, NY}
}

@inproceedings{yang2025gated,
  title={Gated delta networks: Improving mamba2 with delta rule},
  author={Yang, Songlin and Kautz, Jan and Hatamizadeh, Ali},
  booktitle={International Conference on Learning Representations},
  volume={2025},
  pages={29687--29707},
  year={2025}
}

@article{cao2026qwen3,
  title={Qwen3-coder-next technical report},
  author={Cao, Ruisheng and Chen, Mouxiang and Chen, Jiawei and Cui, Zeyu and Feng, Yunlong and Hui, Binyuan and Jing, Yuheng and Li, Kaixin and Li, Mingze and Lin, Junyang and others},
  journal={arXiv preprint arXiv:2603.00729},
  year={2026}
}

@article{hatamizadeh2026gated,
  title={Gated DeltaNet-2: Decoupling erase and write in linear attention},
  author={Hatamizadeh, Ali and Choi, Yejin and Kautz, Jan},
  journal={arXiv preprint arXiv:2605.22791},
  year={2026}
}

@article{gu2023mamba,
  title={Mamba: Linear-time sequence modeling with selective state spaces},
  author={Gu, Albert and Dao, Tri},
  journal={arXiv preprint arXiv:2312.00752},
  year={2023}
}

@inproceedings{lahotimamba,
  title={Mamba-3: Improved Sequence Modeling using State Space Principles},
  author={Lahoti, Aakash and Li, Kevin and Chen, Berlin and Wang, Caitlin and Bick, Aviv and Kolter, J Zico and Dao, Tri and Gu, Albert},
  booktitle={The Fourteenth International Conference on Learning Representations},
  year={2026}
}

@article{auer2026tirex,
  title={Tirex: Zero-shot forecasting across long and short horizons with enhanced in-context learning},
  author={Auer, Andreas and Podest, Patrick and Klotz, Daniel and B{\"o}ck, Sebastian and Klambauer, G{\"u}nter and Hochreiter, Sepp},
  journal={Advances in Neural Information Processing Systems},
  volume={38},
  pages={57529--57580},
  year={2026}
}

@article{podest2026tirex,
  title={TiRex-2: Generalizing TiRex to Multivariate Data and Streaming},
  author={Podest, Patrick and Pichler, Marco and B{\"u}rger, Elias and Z{\'o}lyomi, Levente and Voggenberger, Bernhard and Berghammer, Wilhelm and Klotz, Daniel and B{\"o}ck, Sebastian and Klambauer, G{\"u}nter and Hochreiter, Sepp},
  journal={arXiv preprint arXiv:2607.01204},
  year={2026}
}

@article{huo2026periodic,
  title={Periodic RoPE for Infinite Context LLMs},
  author={Huo, Simin},
  journal={arXiv preprint arXiv:2605.27980},
  year={2026}
}

@article{huang2026mdn,
  title={MDN: Parallelizing Stepwise Momentum for Delta Linear Attention},
  author={Huang, Yulong and Liu, Xiang and Huang, Hongxiang and Lin, Xiaopeng and Liu, Zunchang and Chu, Xiaowen and Xie, Zeke and Cheng, Bojun},
  journal={arXiv preprint arXiv:2605.05838},
  year={2026}
}

@article{yang2024parallelizing,
  title={Parallelizing linear transformers with the delta rule over sequence length},
  author={Yang, Songlin and Wang, Bailin and Zhang, Yu and Shen, Yikang and Kim, Yoon},
  journal={Advances in neural information processing systems},
  volume={37},
  pages={115491--115522},
  year={2024}
}

@article{team2026kimi,
  title={Kimi K3: Open Frontier Intelligence},
  author={Team, Kimi and Bai, Tongtong and Bai, Yifan and Bao, Yiping and Cai, Jianfeng and Cai, Xinyuan and Cao, Peizhou and Cao, Yuxuan and Chai, Ziwei and Charles, Y and others},
  journal={arXiv preprint arXiv:2607.24653},
  year={2026}
}

@inproceedings{ben2025decimamba,
  title={Decimamba: Exploring the length extrapolation potential of mamba},
  author={Ben-Kish, Assaf and Zimerman, Itamar and Abu-Hussein, Shady and Cohen, Nadav and Globerson, Amir and Wolf, Lior and Giryes, Raja},
  booktitle={International Conference on Learning Representations},
  volume={2025},
  pages={101148--101170},
  year={2025}
}

\appendix

\section{Training and Evaluation Details}

We release the kernel implementation of State Neutralization for RWKV-7 at \url{https://github.com/pass-lin/rwkv_ops}. 

All mathematical reasoning results are scored using the \texttt{math\_verify} library.

The training setup is as follows. Both the baseline and SANE models are initialized from the same pretrained checkpoint, RWKV-7-0.4B-G1D. The backbone is loaded via KerasHub with the JAX backend. We use mixed bfloat16 precision throughout training. The optimizer is moonlight version Muon\cite{liu2025muon} with a cosine learning rate schedule. The initial learning rate is 1e-7, which warms up over the first 3\% of total steps to a peak of 5e-5, then decays with cosine annealing to the final step. Weight decay is set to 0.1. Gradient clipping is applied with a global clip norm of 1.0. The Muon-specific hyperparameters are: adam\_lr\_ratio=1, ns\_steps=5, rms\_rate=0.2. Certain parameters are excluded from weight decay, including all token-shift weights , biases, normalization parameters (norm, ln), LoRA parameters, the alpha parameter, and the tau parameter. The embedding layer, output head, and their associated weights are also excluded from weight decay. These parameters excluded from weight decay are trained with AdamW instead of Muon. Likewise, following the general setup for Muon, the input embedding and output head are also trained with AdamW.

The training data consists of approximately 2 million sequences of length 4097, loaded from a preprocessed numpy file. We truncate the dataset to a multiple of the batch size. The batch size is 48. The input is constructed by taking the first 4096 tokens of each sequence as the input token IDs and the last 4096 tokens as the target token IDs. The padding mask is derived from the input token IDs (non-zero positions are valid). The sample weight for the loss is derived from the target token IDs (non-zero positions are counted in the loss).

 The chunk size for State Neutralization is 16, matching the native RWKV-7 chunk size. The SANE module is applied at chunk boundaries during both forward and backward passes. The training runs for 1 epoch. We train and infer our models using the Keras 3 + JAX framework.

For short-context mathematical reasoning evaluation, we use sampling-based decoding with top-p=0.8, top-k=20, and temperature=1.0. We generate 8 samples per problem and take the majority vote (maj@8) as the final answer. For perplexity evaluation on the synthetic long-context data. The long-context synthetic datasets (\texttt{english\_math\_chat} and \texttt{mixed\_chat}) are constructed by concatenating the training splits of DeepMath-103K and Chinese-DeepSeek-R1-Distill-data-110k. The 100M-token reasoning evaluation feeds the entire \texttt{mixed\_chat} sequence as a prefix before presenting the mathematical test problems.

\end{document}